\documentclass[conference]{IEEEtran}
\usepackage{times}

\usepackage{natbib}
\usepackage{array}
\usepackage{relsize}
\usepackage{multirow}
\usepackage{subfig}
\usepackage{amsmath}
\usepackage{amsfonts, bm}
\usepackage[bookmarks=true]{hyperref}
\usepackage{graphics} 
\usepackage{graphicx}
\usepackage{booktabs}
\usepackage{wrapfig}
\usepackage{enumerate}
\usepackage{rotating}
\usepackage{flushend}

\usepackage{color}

\newlength\savedwidth
\newcommand\whline[1]{\noalign{\global\savedwidth\arrayrulewidth
                               \global\arrayrulewidth #1} %
                      \hline
                      \noalign{\global\arrayrulewidth\savedwidth}}

\renewcommand{\Re}{\mathbb{R}}

\begin{document}

\title{Learning to Place New Objects in a Scene}

\author{

Yun Jiang, Marcus Lim, Changxi Zheng and Ashutosh Saxena\\
Computer Science Department, Cornell University, USA.\\
yunjiang@cs.cornell.edu, mkl65@cornell.edu, \{cxzheng,asaxena\}@cs.cornell.edu 
}



%

\maketitle

\begin{abstract}
    Placing is a necessary skill for a personal robot to have in order to perform tasks such as
    arranging objects in a disorganized room. The object placements should not only be stable but
    also be in their semantically preferred placing areas and orientations.
    This is challenging because an environment can have a large variety of 
    objects and placing areas that may not have been seen by the robot before.
    
        In this paper, we propose a learning approach for placing multiple objects
    in different placing areas in a scene.
    Given point-clouds of the objects and the scene, we design appropriate features
    and use a graphical model to encode various properties,
    such as the stacking of objects, stability, object-area relationship and
    common placing constraints. The inference in our model is an integer
    linear program, which we solve efficiently via an LP relaxation.
    We extensively evaluate our approach on 98 objects from 
    16 categories being placed into 40 areas. Our robotic experiments show
    a success rate of 98\% in placing known objects and 82\% in placing new objects stably.
    We use our method on our robots for performing tasks such as loading
    several dish-racks, a bookshelf and a fridge with multiple items.\footnote{Parts of this work were described in \cite{JiangPlacing_workshop,JiangPlacing}.}

\end{abstract}


\IEEEpeerreviewmaketitle

\section{Introduction}
In order to autonomously perform common daily tasks such as setting up a dinner table,
arranging a living room or organizing a closet, a personal robot should be
able to figure out where and how to place objects. 
However, this is particularly challenging because there can potentially be a
wide range of objects and placing environments. Some of them may not 
have been seen by the robot before.
For example, to tidy a disorganized house, a robot needs
to decide \textit{where} the best place for an object 
is (e.g., books should be
placed on a shelf or a table and plates are better inserted in a dish-rack),
and \textit{how} to place the objects in an area 
(e.g. clothes can be hung in a closet  and 
wine glasses can be held upside down on a stemware holder).
In addition, limited space, such as in a cabinet, raises another problem of how to
\textit{stack} various objects together for efficient storage.  Determining
such a placing strategy, albeit rather natural or even trivial to (most)
people, is quite a challenge for a robot.

\begin{figure}
\begin{center}
\includegraphics[width=1.0\linewidth]{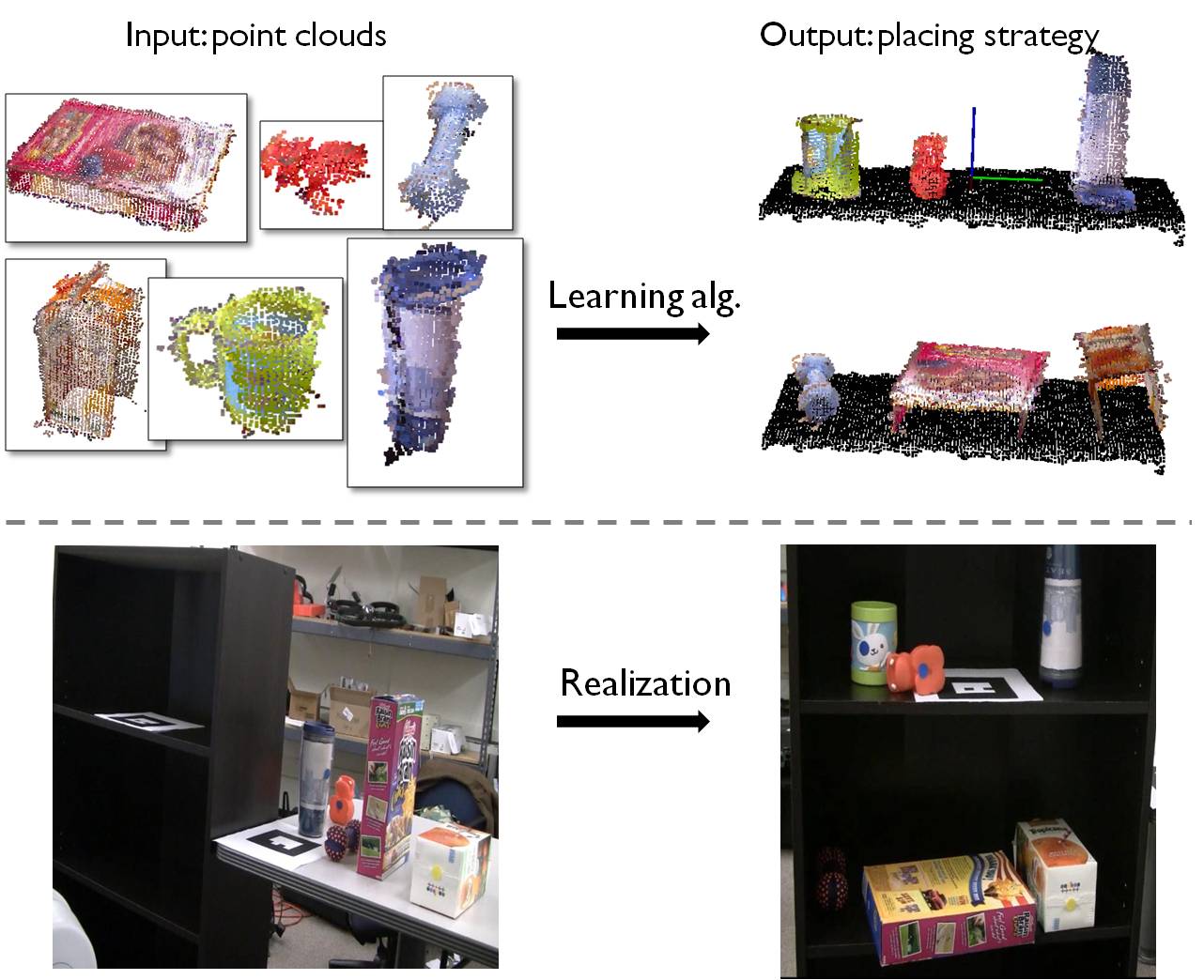}
\end{center}
\caption{\small{An example task of placing items on a bookshelf. 
Given the point-clouds of the bookshelf and six objects to be placed (shown in top-left
part of the figure), our
learning algorithm finds out the best placing strategy, specified by the
location and orientation of every object (shown in top-right). 
Following this inferred strategy, the robot places each object accordingly. The bottom
part of the figure shows the scene before and after placing.
Note that in some cases,
the placing environment can be quite complex (e.g, see Fig.~\ref{fig:exp2}).
}
}
\label{fig:placing_def}
\end{figure}

In this paper, we consider multiple objects and placing areas represented by possibly incomplete 
and noisy point-clouds. Our goal is to find proper placing strategies to place the 
objects into the areas. A placing \emph{strategy} of
an object is described by a preferred placing area for the 
object and a 3D location and orientation to place it in that area.
As an example, Fig.~\ref{fig:placing_def} shows one possible strategy to place six different types of objects onto
a bookshelf. 
In practice, the following criteria should be considered.

\noindent
\textbf{Stability}: Objects should be placed stably in order to avoid falling.
For example, some objects can be stably placed upright on flat surfaces,\footnote{Even knowing 
the ``upright" orientation for an arbitrary object is a
non-trivial task \citep{siggraph_upright}.} plates can be placed stably in
different orientations in a dish-rack, and a pen can be placed horizontally on a table but
vertically 
in a pen-holder.

\noindent
\textbf{Semantic preference}: 
A placement should follow common human preferences in placing.
For instance, shoes should be placed on the
ground but not on a dinner plate, even though both areas have geometrically similar
flat surfaces. Therefore, a robot should be able to distinguish the areas semantically,
and make a decision based on common human practice.


\noindent
\textbf{Stacking}: A placement should consider possible stacking of objects such as piling up dinner plates.
However, this raises more challenges for a robot because it has to decide
which objects can be stacked together semantically. 
For example, it is a bad idea to stack a dinner plate
on top of a cell phone rather than another dinner plate. In addition, the robot
has to decide the order of stacking in a dynamically changing environment,
since previously placed objects can change the structure of placing areas for 
objects placed later.

In addition to the difficulties introduced by these criteria, perceiving the
3D geometry of objects and their placing environments is nontrivial as well. In
this paper, we use a depth camera mounted on a robot to perceive the 3D geometries
as point-clouds. In practice, the perceived point-clouds can be noisy and
incomplete (see Fig.~\ref{fig:placing_def}), requiring the robot to be able to 
infer placements with only partial and noisy geometric information.


In this paper, we address these challenges using a learning-based approach.
We encode human preferences about placements as well as
the geometric relationship between objects and their placing environments by
designing appropriate features. 
We then propose a graphical model that has two substructures to capture the stability and the semantic
preferences respectively. The model also incorporates stacking and constraints that keeps the placing
strategy physically feasible. 
We use max-margin learning for estimating the parameters in our graphical model.
The learned model is then used to score the potential placing strategies. 
Given a placing task, although inferring the best strategy (with the highest score) is 
provably NP-complete,  we express the inference as an integer linear programming (ILP) 
problem which is then solved efficiently using an linear programming (LP) relaxation.


To extensively test our approach, 
we constructed a large placing database composed of 98 household objects from 16 different
categories and 40 placing areas from various scenes.
Experiments ranged from placing a single object in challenging situations
to complete-scene placing where we placed up to 50 objects in real offices and apartments.
In the end-to-end test of placing multiple objects in different scenes, 
our algorithm significantly improves the performance---on metrics of stability, semantic 
correctness and overall impressions on human subjects---as compared
to the best baseline.
Quantitatively, we achieve an average accuracy of 83\% for stability and
82\% for choosing a correct placing area for single-object placements.
Finally, we tested our algorithm on two different robots on several placing tasks. 
We then applied our algorithm to several practical placing scenarios,
such as loading multiple items in dish-racks, loading a fridge, placing objects
on a bookshelf and cleaning a disorganized room.  We have also made the code
and data available online at: 
\texttt{http://pr.cs.cornell.edu/placingobjects}
	 
Our contributions in this paper are as follows:
\begin{itemize}    
\item There has been little work on robotic placing, compared to other manipulation
tasks such as grasping. To the best of our knowledge, this is the first
work on placing objects in a complex scene. In particular, our scenes are 
composed of different placing areas 
such as several dish-racks, a table, a hanging rod in a closet, etc.
\item We propose a learning algorithm that takes object stability 
as well as semantic preferences into consideration. 
\item Our algorithm can handle objects perceived as noisy and incomplete
point-clouds. The objects may not have been seen by the robot before.
\item We consider stacking when placing multiple objects in a scene.
\item We test our algorithm extensively on a variety of placing tasks, both on
offline real data and on robots. 
\end{itemize}

This paper is organized as follows. We start with a review of the related work in
Section~\ref{sec:relatedwork}. We describe our robot platforms in Section~\ref{sec:platforms}. 
We then formulate the placing problem 
in a machine learning framework in Section~\ref{sec:definition}. We present our learning
algorithm for single-object placements in Section~\ref{sec:alg_single} and the corresponding 
experiments in Section~\ref{sec:exp_single}. 
We give the algorithm and the experiments for multiple-object placements  
in Section~\ref{sec:alg_multi} and Section~\ref{sec:exp_multi} respectively. 
Finally, we conclude the paper in Section~\ref{sec:conclusion}.

\section{Related Work}
\label{sec:relatedwork}

While there has been significant previous work on grasping objects
\citep[e.g.,][]{review_grasping, GraspIt,Miller2003, SaxenaGraspingNIPS,SaxenaIJRR,SaxenaAAAI, berenson2009grasp, rao-grasping, hsiao_haptic_grasping, brook:grasping:2011, QuocLeRank,
jiangICRA,dogar2011framework, rosalesglobal,JiangICRA12},
there is little work on object placement, and it is restricted to placing
objects on flat horizontal surfaces. 
For example, \cite{kemp-placing} recently developed a learning algorithm to
detect clutter-free `flat' areas where an object can be placed. 
Unfortunately, there are many non-flat placing areas where this method would not apply. 
Even if placing only on flat surfaces, one needs to decide the upright or the current orientation of a given object, which is a challenging task.
For example, \cite{siggraph_upright} proposed
several geometric features to learn the upright orientation
from an object's 3D model and \cite{saxena_orientation} predicted
the orientation of an object given its 2D image. Recently, \cite{Glover-RSS-11}
used a database of models to estimate the pose of objects with partial point-clouds. 
Our work is different and complementary to these studies: we generalize
placing environment from flat surfaces to more complex
ones, and desired configurations are extended from upright
to all other possible orientations that can make the best use
of the placing area.
Furthermore, we consider:
1) placing objects in scenes 
comprising
a wide-variety of 
placing areas, such as dish-racks, stemware holders, cabinets and hanging rods in closets;
2) placing multiple objects;
3) placing objects in semantically preferred locations (i.e., how to choose a proper 
placing context for an object).

For robotic placing, one component is to plan and control the arm to place the
objects without knocking them down.
\cite{edsinger2007manipulation}
considered placing objects on  a flat shelf, but their focus was to use passive compliance and
force control to  gently place the object on the shelf. 
Planning and rule-based approaches have been used 
to move objects around. For example, \cite{lozano2002task} 
proposed a task-level (in contrast with motion-level) planning system and tested it on picking and placing objects on a table. 
\cite{sugie2002placing} used rule-based planning in order to push objects on
a table surface. However, most of these approaches assume known
full 3D models of the objects, consider only flat surfaces,
and do not model semantic preferences in placements.

Placing multiple objects also
requires planning and high-level reasoning about the order of execution.
\cite{berenson2009grasp} coupled planning and grasping in cluttered scenes.
They utilized an `environment clearance score' in determining which object to grasp first. 
\cite{Toussaint2010} integrated control, planning, grasping and reasoning in the
`blocks-world' application in which table-top objects were rearranged into several stacks by a robot.
How to arrange objects efficiently is also related to the classic bin packing
problem~\citep{coffman1996approximation} which can be approached as integer 
linear programming~\citep{fisher1981lagrangian},
constraint satisfaction problem~\citep{pisinger2007using} or tabu search~\citep{lodi2002heuristic}.
These studies focus on generic planning problems and are complementary to ours.


Contextual cues \citep[e.g.,][]{context_torralba} have  proven helpful in many vision 
applications. 
For example, 
using estimated 3D geometric properties from images can be useful for object detection
\citep{saxena-ijcv-2008,saxena-pami2009,context_ccm,context_feccm,spatial_layout_cluttered_rooms,divvala2009empirical}.
In \cite{xiong2010using} and \cite{AnandKoppulaNIPS}, contextual information was 
employed to estimate semantic labels in 3D point-clouds of indoor environments.  
\cite{fefeili_retrieval} used object context for object retrieval.
\cite{fisher2010context} and \cite{fishercharacterizing} designed
a context-based search engine using geometric cues and
spatial relationships to find the proper object for a given scene. Unlike our work, 
their goal was only to retrieve the object but not to place it afterwards.
While these works address different problems, 
our work that captures the semantic preferences in placing is motivated by them.

Object categorization is also related to our work as objects from same category often share similar placing patterns. 
Categorization in 2D images is a well-studied computer vision problem. Early work 
\citep[e.g.,][]{winn2005object, leibe2004combined, fergus2003object, berg2005shape} 
tried to solve shape and orientation variability, limited by a single viewpoint. 
Motivated by this limitation, multi-view images were considered for categorizing
3D generic object by connecting 2D features
\citep{thomas2006towards,savarese20073d}. When 3D models were available, some work categorized
objects based on 3D features instead, e.g., using synthetic 3D data to extract pose and class 
discriminant features \citep{liebelt2008viewpoint},
and using features such as spin images \citep{johnson1999using} and point feature histograms \citep{rusu2008learning} for 3D recognition. 
Instead of images or 3D models, \cite{laisparse} proposed a learning algorithm 
for categorization using both RGB
and multi-view point-cloud. These works  are complementary to ours in that we 
do not explicitly categorize the object before placing, but
knowing the object category could potentially help in placing.

Most learning algorithms require good features as input, and often these features
are hand-designed for the particular tasks \citep{siggraph_upright,SaxenaAAAI}.
There have also been some previous works on high-dimensional 3D features \citep{Ho2010,rusu2008learning,johnson1999using,liebelt2008viewpoint,SaxenaAAAI} but
they do not directly apply  to our problem. 
There is also a large body of work on automatic feature selection \citep[see][]{dy2004feature, liu2005toward,Jetchev2011},
which could potentially improve the performance of our algorithm.



In the area of robotic manipulation, a wide range of problems 
have been studied so far,
 such as folding towels and clothes~\citep{maitin2010cloth}, opening
 doors~\citep{Kemp2010pulling,klingbeil_door}, inferring 3D articulated objects \citep{sturm20103d,katz2008manipulating}
 and so on. 
However, they address different manipulation tasks and do not apply
to the placing problem  we consider in this paper. 
Our work is the first one to consider object placements in complex scenes.

\section{Robot Platforms and System}
\label{sec:platforms}

\begin{figure}
\begin{center}
\includegraphics[width=1\linewidth]{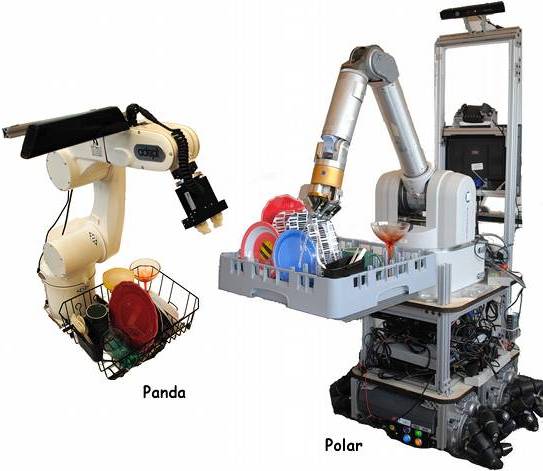}
\end{center}
\caption{\small{Our two robotic platforms used for testing our algorithm: PANDA (PersonAl
Non-Deterministic Arm, on left) is an Adept arm with a parallel-plate gripper,
mounted with a Kinect sensor. POLAR (PersOnaL Assistant Robot,
on right) is a 7-DOF Barrett arm mounted on a Segway Omni base, with a three-fingered hand
and a Kinect sensor on top.}}
\label{fig:pr_robots}
\end{figure}
In this section, we describe our robot platforms and details of the system.

Our primary sensor is a depth camera that gives an RGB image together with the depth value at each
pixel. This could be converted into a colored point-cloud as in Fig.~\ref{fig:placing_def}.
In our experiments, we used a Microsoft Kinect sensor 
which has a resolution of 640x480 for the depth image and an operation range of 0.8m to
3.5m. 

Our PANDA (PersonAl Non-Deterministic Arm) robot (Fig.~\ref{fig:pr_robots} left) is a 6-DOF
Adept Viper s850 arm equipped with a parallel-plate gripper and a Kinect sensor. 
The arm, together with the gripper, has a
reach of 105cm. The arm has a repeatability of 0.03mm in
XYZ positioning, but the estimated repeatability with the gripper is 
0.1mm. The Kinect sensor-arm calibration was accurate up to an average of 3mm.
The arm weighs 29kg and has a rated payload of 2.5kg, but our gripper can only
hold up to 0.5kg. The Adept Viper is an industrial arm and
has no force or tactile feedback, so even a slight error in
positioning can result in a failure to place. 

Our POLAR (PersOnaL Assistant Robot) robot (Fig.~\ref{fig:pr_robots} right) is 
equipped with a 7-DOF WAM
arm (by Barrett Technologies) and a three-fingered hand mounted on a Segway RMP 50 Omni base. 
A Kinect sensor is mounted on top. The arm has a positioning accuracy of $\pm$0.6mm.
It has a reach of 1m and a payload of 3kg. The hand does not have tactile
feedback. The mobile base can be controlled by speed and has a
positioning accuracy of only about $10$cm. 

Our system comprises three steps: perception, inference and
realization. We first use the sensor to capture the point-cloud of the object and
the scene. We then apply the learning algorithm to find good placements. In the last step, given the 3D location and
orientation of the placing point, the robot uses path planning to move the
object to the destination and releases it in the end.

In a more complex task involving placing multiple objects (such as organizing a room in
Section~\ref{sec:exp_multi_robotic}), the robot needs to move to the
object, pick up the object, move to the designated
area, find out the location and orientation for placing, and place it
accordingly. 
We start with a 3D model of the room. The initial locations of the objects
to be placed are given and their grasps are defined in advance. We use open-source softwares such as
ROS~\citep{ROS} and OpenRAVE~\citep{openrave} for path planning and obstacle avoidance,
and use AR-markers~\citep{armarker}
for robot localization.
Our goal in this paper is to develop an algorithm for estimating good placements given the
point-clouds of the objects and the scene as input.

\section{Problem Formulation}\label{sec:definition}
In this section, we formulate the problem of how to predict good placements 
in a placing task. 



Specifically, our goal is to place a set of objects in a scene that 
can contain several potential placing areas.
Both the objects and the placing areas are perceived as point-clouds that can be noisy and
incomplete.  A placement of an
object is specified by 1) a 3D location describing at which 3D position in the scene the
object is placed, and 2) a 3D rotation describing the orientation of the object
when it is placed.  In the following, we first consider the problem of single-object placements.
We will then extend it to multiple-object placements by adding semantic features
and modifying the algorithm to handle multiple objects. The learning algorithm for 
single-object placements is only a special case of the algorithm for multiple-object 
placements.



\subsection{Single-Object Placements}
Here, our goal is to place an object stably in a designated placing area
in the scene.  We consider stability and  preferences in orientation, 
but would not consider the semantic preferences about which placing area to choose. 
Specifically, the goal is to infer the 3D location $\ell$ (in a placing area $E$)
and 3D orientation $c$ in which to place the object $O$.

\begin{figure}
\centering
\includegraphics[width=1\linewidth]{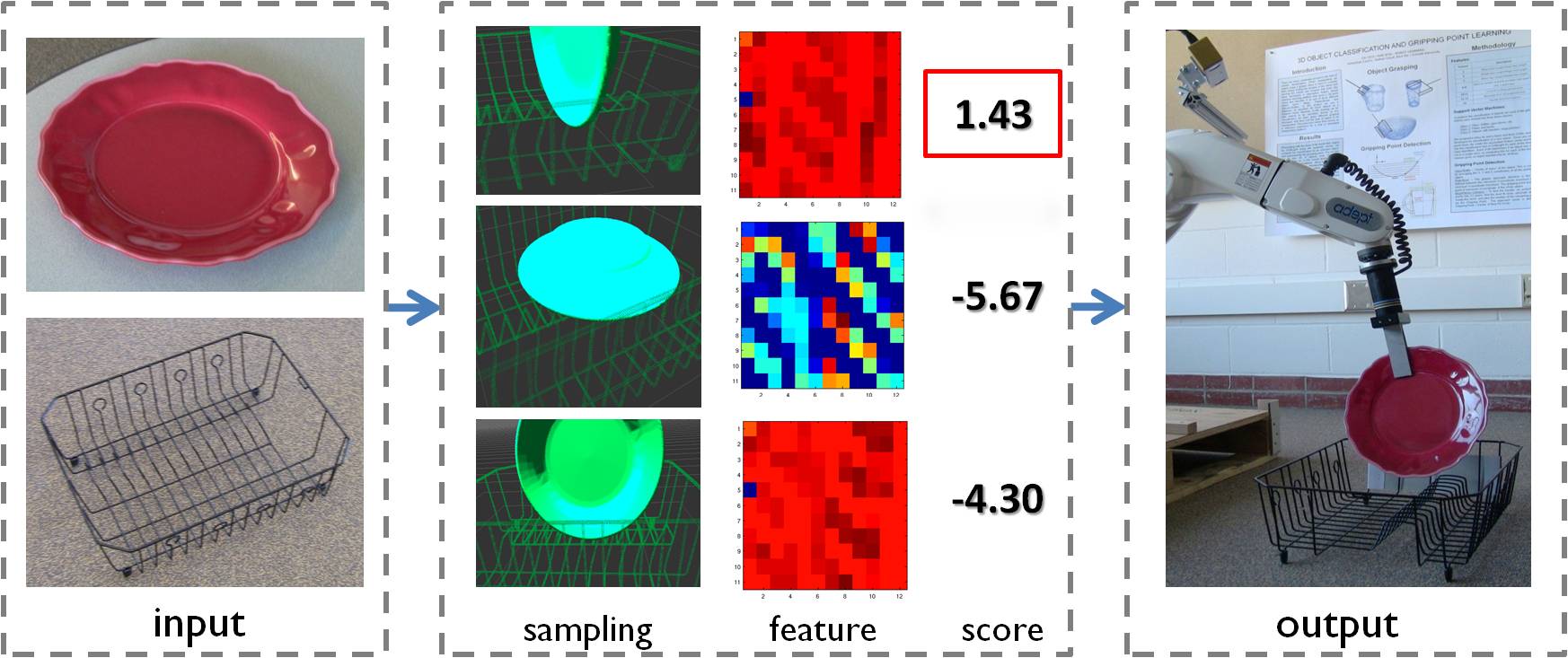}
\caption{{\small Our formulation of single-object placements as a learning problem.
Steps from left to right: 1) we are given an object to be placed
and a placing area, in the form of point-clouds; 2) we first sample possible
placements, extract features for each sample, and then use our learned model
to compute a score for each sample. The higher the score is, the more likely
it is to be a good placement; 3) the robot plans a path to the predicted
placement and follows it to realize the placing.
}} 
\label{fig:single_formulation}
\end{figure}

We illustrate the problem formulation in Fig.~\ref{fig:single_formulation}.
We are given an object $O$ and a placing area $E$ in the form of point-clouds.
Given the input, we first sample a set of
possible placements, compute relevant features for each, and then use the
learned model to compute a score for each candidate. The highest-score
placement is then selected to be executed by the robot. 
We describe our features and the learning algorithm in Section~\ref{sec:alg_single}.


\subsection{Multiple-Object Placements}\label{sec:def_multi}
In addition to finding a proper location and orientation to place the object, we
often need to decide which placing area in the scene is semantically suitable
for placing the object in. When multiple objects are involved, we also need
to consider stacking 
while placing them.


As an example, consider organizing a kitchen (see Fig.~\ref{fig:kitchen_example}).
In such a case,
our goal is to place a set of given objects
into several placing areas. One can place objects in two ways: directly on an
existing area in the environment (e.g., saucepans on the bottom drawer), or stacking 
one on top of another (e.g., bowls piled up in the cabinet). 

Formally, the input of this problem is $n$ objects $\mathcal O=\{O_1, \ldots,O_n\}$ and
$m$ placing areas (also called environments) $\mathcal E=\{E_1,\ldots,E_m\}$, all of
which are represented by point-clouds. The output will be a placing strategy depicting the final
layout of the scene after placing. It is specified by the pose of every object, which
includes the configuration (i.e., 3D orientation) $c_i$, the placing area (or
another object when stacking) selected, and the relative 3D location $\ell_i$
w.r.t. this area. We propose a graphical model to represent the placing
strategy, where we associate every possible strategy with a potential function. 
Finding the best placing strategy is then equivalent to maximizing the potential function.
Details are in Section~\ref{sec:alg_multi}.

\begin{figure*}[thb!] \centering
    \subfloat[supporting contacts] {\includegraphics[width=0.33\columnwidth]{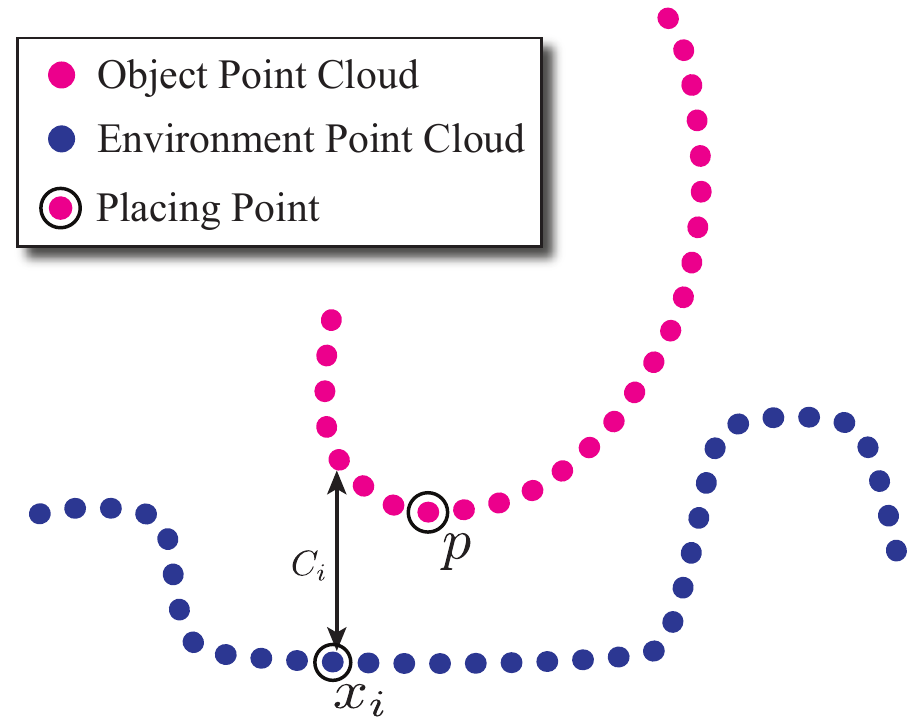}\label{fig:features_a}}
    \qquad
    \subfloat[caging (top view)] {\includegraphics[width=0.33\columnwidth]{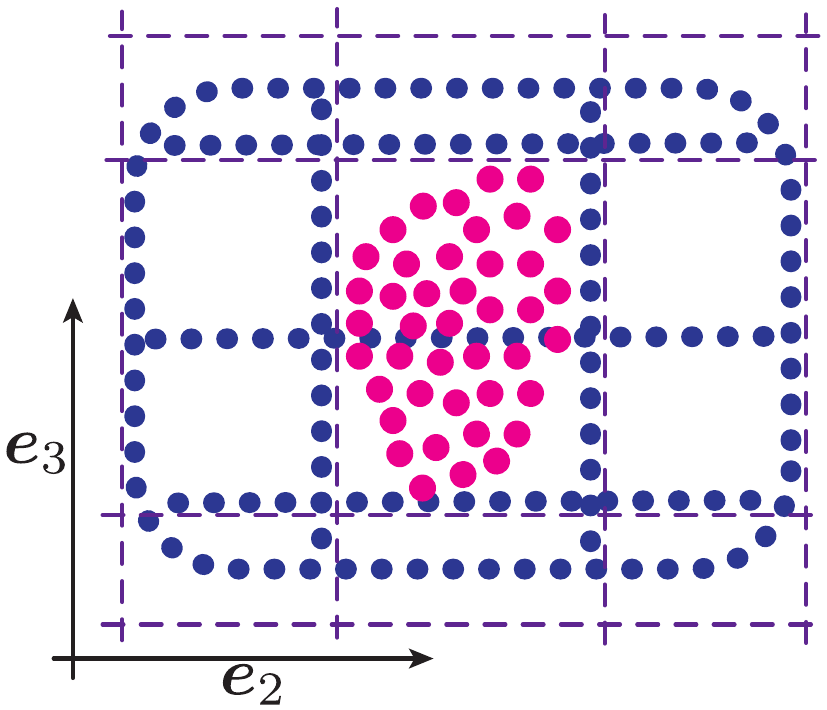}\label{fig:features_b}}
    \qquad
    \subfloat[caging (side view)] {\includegraphics[width=0.34\columnwidth]{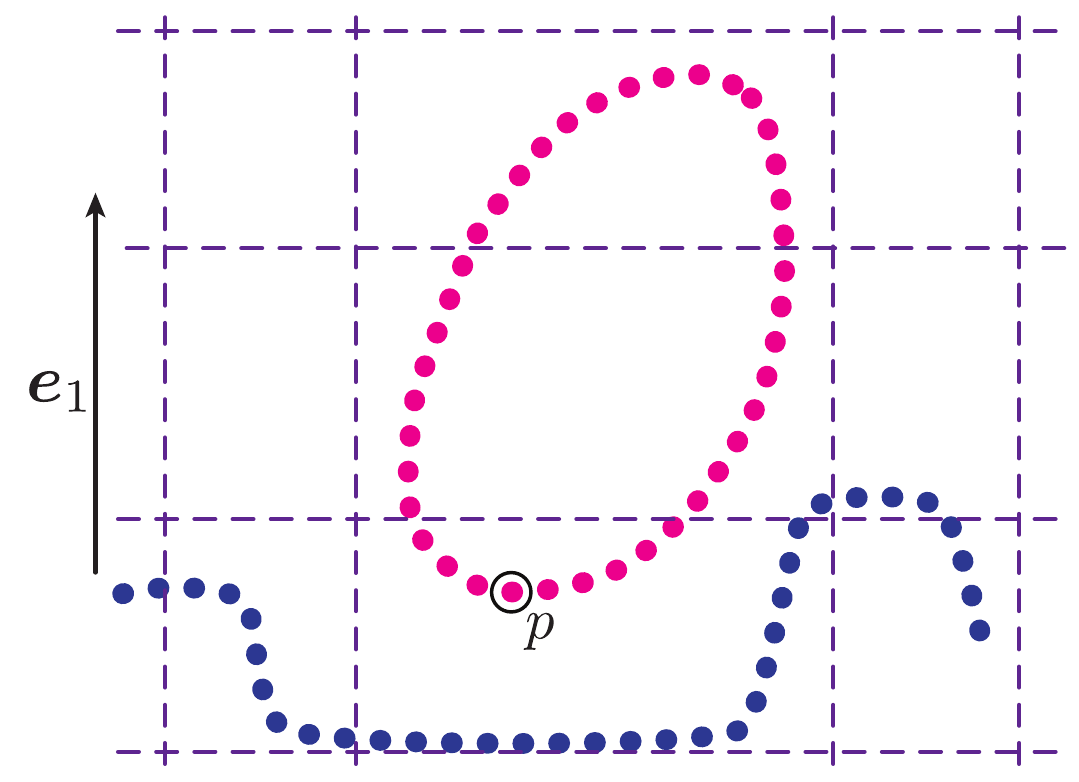}\label{fig:features_c}}
    \qquad
    \subfloat[histogram (top view)] {\includegraphics[width=0.33\columnwidth]{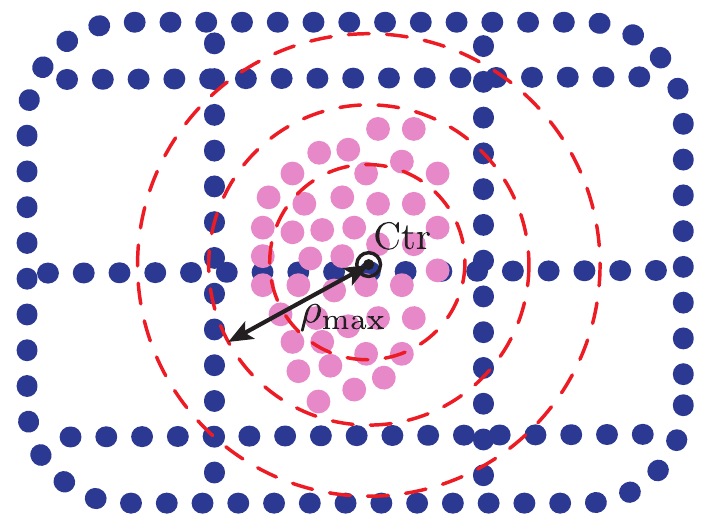}\label{fig:features_d}}
    \qquad
    \subfloat[histogram (side view)] {\includegraphics[width=0.32\columnwidth]{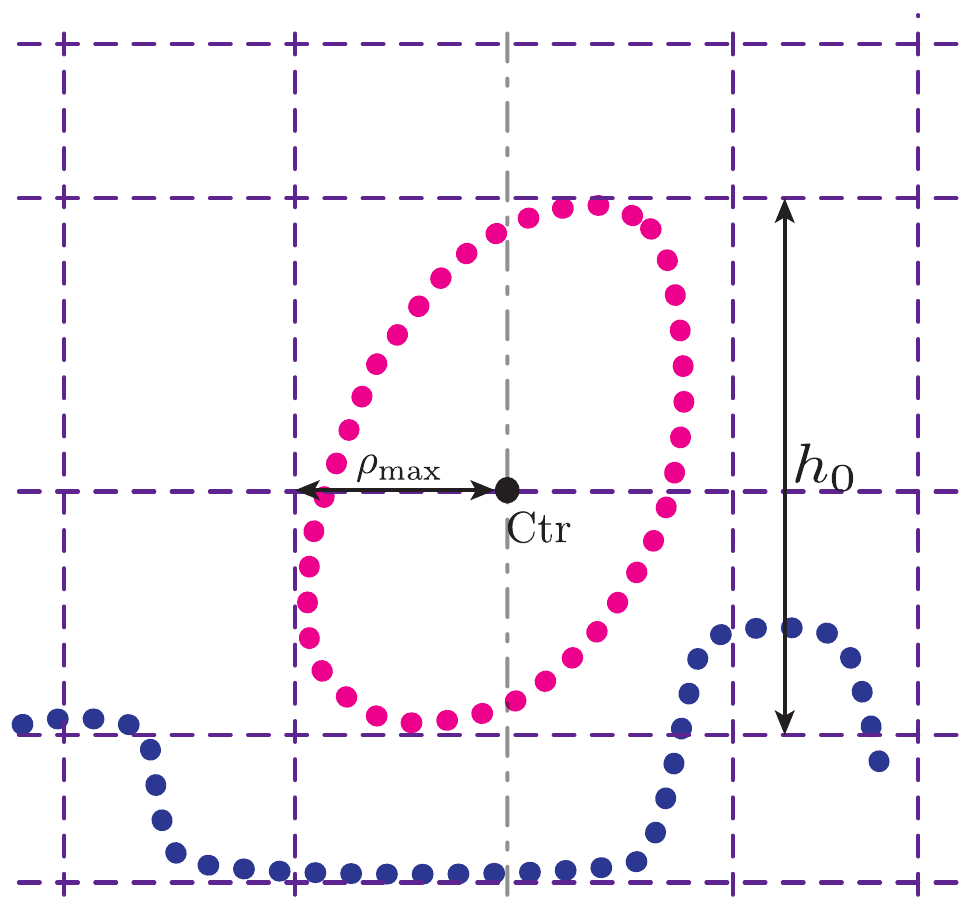}\label{fig:features_e}}
    \caption{{\small Illustration of features in our learning algorithm for single-object placements.
        These features are designed to capture the stability and preferred orientations in a good
            placement.}}
    \label{fig:features}
\end{figure*}

\section{Algorithm for Single-Object Placements}
\label{sec:alg_single}


In order to identify good placements, we first need to design features that 
indicate good placements across various objects and placing areas. 
We then use a max-margin learning algorithm to
learn a function that maps a placement, represented by its features, to a
placing score. In testing, we first randomly sample some placements, then
use the function to find the highest-score candidate as our best placement.

\subsection{Features}\label{sec:feature_single}

The features used in our learning algorithm are designed to capture 
the following two properties:

\begin{itemize}
\item \textbf{Supports and Stability.}
The object should stay still after placing. Ideally, it should also be 
able to withstand small perturbations. 

\item \textbf{Preferred Orientation.}
A good placement should have semantically preferred orientation as well. 
For example, plates should be inserted into a dish-rack vertically
and glasses should be held upside down on a stemware holder. 
\end{itemize}

An important property of the stability features is invariance
under translation and rotation (about the gravity, i.e., Z-axis). This is because
as long as the relative configuration of the object and the placing area
remains same, the features should not change. Most of our features will follow
this property.

We group the features into three categories.
In the following description, we use $\mathsf{O}'$ to denote the point-cloud of the object $O$
after being placed, and use $\mathsf{B}$ to denote the point-cloud of a placing area $B$. 
Let $\bm{p}_o$ be the 3D coordinate of a point $o\in \mathsf{O}'$ from the object, and $\bm{x}_t$ be the coordinate of a
point $t\in \mathsf{B}$ from the placing area. 

\smallskip
\noindent\textbf{Supporting Contacts:}
Intuitively, an object is placed stably when it is supported by a wide spread of contacts. Hence, we
compute features that reflect the distribution of the supporting contacts. In particular, we choose the
top $5\%$ points in the placing area closest to the object (measured by the vertical distance, $c_i$
shown in Fig.~\ref{fig:features_a}) at the placing point. 
Suppose the $k$ points are
$\bm x_1,\ldots,\bm x_k$. We quantify the set of these points by 8 features: 
\begin{enumerate}
\item Falling distance $\min_{i=1\dots k} c_i$.
\item Variance in XY-plane and Z-axis respectively,  $\frac{1}{k}\sum_{i=1}^k (\bm x'_i - \bar{\bm
x}')^2$, where $\bm x'_i$ is the projection of $\bm x_i$ and $\bar{\bm x}' = \frac{1}{k}\sum_{i=1}^k
\bm x'_i$.
\item Eigenvalues and ratios. We compute the three Eigenvalues ($\lambda_1\geq \lambda_2\geq \lambda_3$) of the covariance matrix of these $k$
points. Then we use them along with the ratios
$\lambda_2/\lambda_1$ and $\lambda_3/\lambda_2$ as the features. 
\end{enumerate}

Another common physics-based criterion is the center of mass (COM)
of the placed object should be inside of (or close to) the region enclosed by contacts.
So we calculate the distance from the centroid of $\mathsf{O}'$ to the nearest boundary of the 2D convex hull formed by contacts
projected to XY-plane, $\mathcal H_{con}$. We also compute the projected convex
hull of the whole object, $\mathcal H_{obj}$. The area ratio of these two
polygons $S_{\mathcal H_{con}}/S_{\mathcal H_{obj}}$ is included as 
another feature.

Two more features representing the percentage of the object points below or above the 
placing area are used to capture the relative location.

\smallskip
\noindent\textbf{Caging:}
There are some placements where the object would not be strictly immovable but is well
confined within the placing area. A pen being placed upright in a pen-holder is one example. 
While this kind of placement has only a few supports from the pen-holder and may
move under perturbations, it is still considered a good one. We call this effect
`gravity caging.'\footnote{This idea is motivated by previous works
 on force closure~\citep{nguyen1986,ponce1993computing} and caging
 grasps~\citep{diankov2008manipulation}.}

We capture this by partitioning the point-cloud of the
environment and computing a battery of features for each zone. In detail,  
we divide the space around the object into $3\times3\times3$ zones. The whole
divided space is the axis-aligned bounding box of the object scaled by $1.6$,
and the dimensions of the center zone are $1.05$ times those of the bounding box
(Fig.~\ref{fig:features_b} and \ref{fig:features_c}). The point-cloud of the
placing area is partitioned into these zones labelled by $\Psi_{ijk},\, i,j,k \in \{1,2,3\}$,
where $i$ indexes the vertical direction $\bm{e}_1$, and $j$ and $k$ index the
other two orthogonal directions, $\bm{e}_2$ and $\bm{e}_3$, on horizontal plane.

From the top view, there are 9 regions (Fig.~\ref{fig:features_b}), each of which
covers three zones in the vertical direction. The maximum height of points in each region is
computed, leading to 9 features.
We also compute the horizontal distance to the object in three vertical levels from four directions
($\pm\bm{e}_2,\pm\bm{e}_3$) (Fig.~\ref{fig:features_c}). In particular, for
each $i=1,2,3$, we compute
\begin{equation}
\begin{split}
d_{i1} &= \min_{ \substack { \bm{x}_t\in\Psi_{i11}\cup\Psi_{i12}\cup\Psi_{i13} \\ \bm{p}_o\in\mathsf{O}'} } {  \bm{e}_2^T(\bm{p}_o - \bm{x}_t)} \\
d_{i2} &= \min_{ \substack { \bm{x}_t\in\Psi_{i31}\cup\Psi_{i32}\cup\Psi_{i33} \\ \bm{p}_o\in\mathsf{O}'} } { -\bm{e}_2^T(\bm{p}_o - \bm{x}_t)} \\
d_{i3} &= \min_{ \substack { \bm{x}_t\in\Psi_{i11}\cup\Psi_{i21}\cup\Psi_{i31} \\ \bm{p}_o\in\mathsf{O}'} } {  \bm{e}_3^T(\bm{p}_o - \bm{x}_t)} \\
d_{i4} &= \min_{ \substack { \bm{x}_t\in\Psi_{i13}\cup\Psi_{i23}\cup\Psi_{i33} \\ \bm{p}_o\in\mathsf{O}'} } { -\bm{e}_3^T(\bm{p}_o - \bm{x}_t)}
\end{split}
\end{equation}
and produce 12 additional features. 

The degree of gravity-caging also depends on the relative height
of the object and the caging placing area. Therefore, we
 compute the histogram of the height of the points surrounding the
object.
In detail, we first define a cylinder centered at the lowest contact point 
with a radius that can just cover $\mathsf{O}'$. Then points of $\mathsf B$ in 
this cylinder are divided into $n_r\times n_\theta$
parts based on 2D polar coordinates. Here, $n_r$ is the number of radial divisions
and $n_\theta$ is the number of divisions in azimuth. 
The vector of the maximum height in each cell (normalized
by the height of the object), $H =
(h_{1,1},\ldots,h_{1,n_\theta},\ldots,h_{n_r,n_\theta})$, is used as additional caging
features. To make $H$ rotation-invariant, the cylinder is always rotated to align polar axis with the highest point, 
so that the maximum value in H is always one of $h_{i, 1},
i=1...n_r$. We set $n_r=4$ and $n_\theta=4$ for single-object placement experiments.

\smallskip
\noindent\textbf{Histogram Features:}
Generally, a placement depends on the geometric shapes of both the
object and the placing area. We compute a representation of the geometry as follows.
We partition the point-cloud of $\mathsf O'$ and of $\mathsf B$ radially and in Z-axis, 
centered at the centroid of $\mathsf O'$.
Suppose the height of the object is $h_O$ and its radius is $\rho_{max}$. The 2D histogram with
$n_z\times n_\rho$ number of bins covers the cylinder with the radius of
$\rho_{max}\cdot n_\rho/(n_\rho-2)$ and the height of $h_O\cdot n_z/(n_z-2)$, illustrated in Fig.~\ref{fig:features_d} and~\ref{fig:features_e}.
In this way, the histogram (number of points in each cell) can capture the global shape of the
object as well as the local shape of the environment around the placing point. 
We also compute the ratio of the two histograms as another set of features, i.e., the number of
points from $\mathsf{O}'$ over the number of points from $\mathsf{B}$ in each cell. The maximum
ratio is fixed to $10$ in practice. For single-object placement experiments, we set $n_z = 4$ and $n_\rho = 8$ and hence have $96$ histogram features.


In total, we generate $145$ features for the single-object placement
experiments: $12$ features for supporting contacts, 
$37$ features for caging, and $96$ for the histogram features.

\subsection{Max-Margin Learning}\label{sec:svm_single}
Under the setting of a supervised learning problem, we are given a dataset of labeled good and bad
placements (see Section~\ref{sec:exp_single_data}), represented by their features. 
Our goal is to learn a function of features that can
determine whether a placement is good or not. As support vector machines (SVMs)~\citep{SVM} have
strong theoretical guarantees in the performance and have been applied to many classification problems, 
we build our learning algorithm based on SVM. 

Let $\phi_i \in \mathbb{R}^p$ be the features of $i^{th}$ instance in the dataset,
and let $y_i \in \{-1,1\}$ represent the label, where $1$ is a good placement 
and $-1$ is not. For $n$ examples in the dataset, the soft-margin SVM learning
problem \citep{SVMlight} is formulated as:
\begin{align}        
\min& \frac{1}{2} \left\|\theta\right\|^2_2+C\sum_{i=1}^n \xi_i \notag\\
\text{s.t. }&  y_i(\theta^T\phi_i-b) \geq 1-\xi_i, \quad  \xi_i\geq 0, \ \forall 1\leq i\leq n
\label{eq:svm_soft}
\end{align}
where $\theta \in \mathbb{R}^p$ are the parameters of the model, and $\xi$ are the slack
variables. This method finds a separating hyperplane that maximizes the margin
between the positive and the negative examples.
Note that our formulation here is a special case of the graphical model for the
multiple-object placements described in Section~\ref{sec:alg_multi}. We also
use max-margin learning to estimate the parameters in both cases.


\begin{figure*}[!t]
\centering
\includegraphics[height=0.083\textheight]{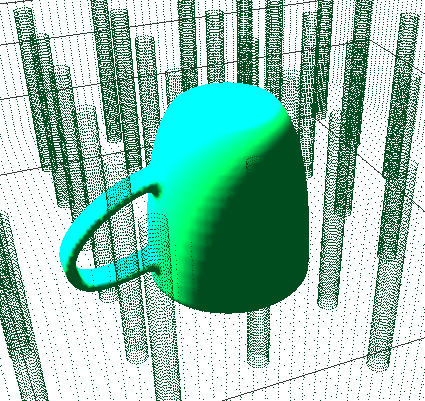}
\includegraphics[height=0.083\textheight]{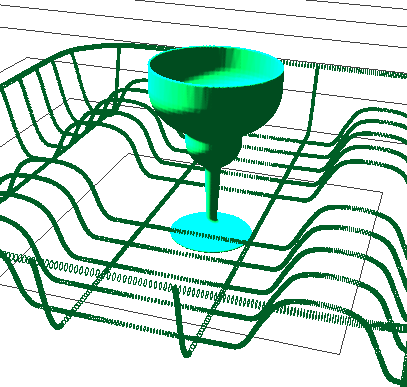}
\includegraphics[height=0.083\textheight]{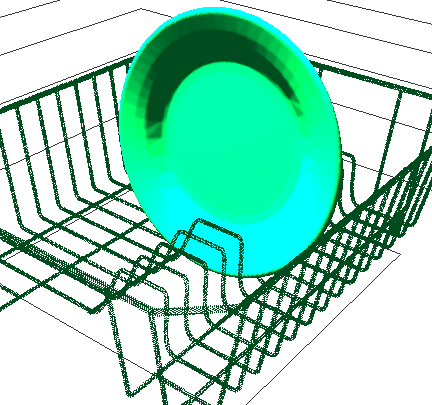}
\includegraphics[height=0.083\textheight]{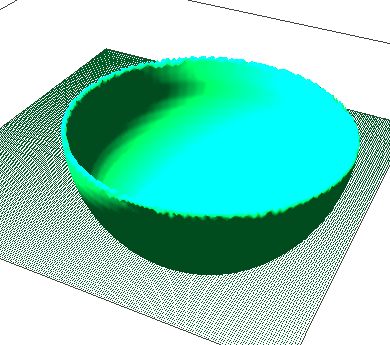}
\includegraphics[height=0.083\textheight]{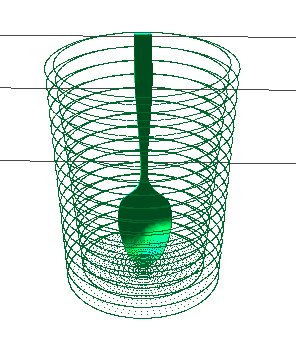}
\includegraphics[height=0.083\textheight]{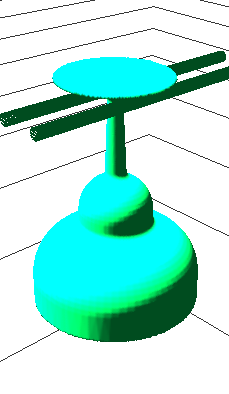}
\includegraphics[height=0.083\textheight]{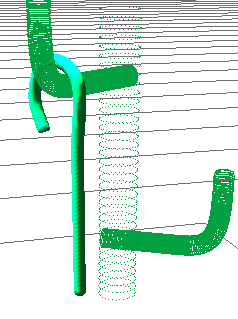}
\includegraphics[height=0.083\textheight]{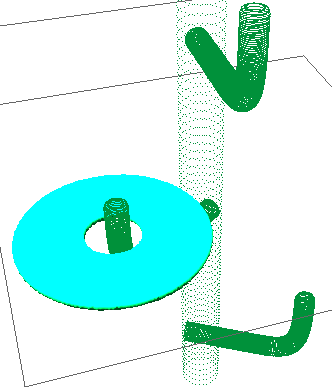}
\includegraphics[height=0.083\textheight]{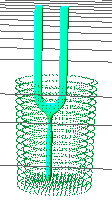}
\caption{{\small Some snapshots from our rigid-body simulator showing different objects
placed in different placing areas. Placing areas from left: rack1, rack2, rack3, flat surface, pen holder,
stemware holder, hook, hook and pen holder. Objects from left: mug, martini glass, plate, bowl, spoon, martini glass, candy cane, disc and tuning fork.
Rigid-body simulation is only used in labeling the training data (Section~\ref{sec:exp_single_data}) and
in first half of the robotic experiments when 3D object models are used
(Section~\ref{sec:exp_single_robotic}).
}}\label{fig:4_scenes}
\end{figure*}

\subsection{Shared-sparsity Max-margin Learning}\label{sec:shared_sparsity_svm}
If we look at the objects and their placements in the environment, we notice
that there is an intrinsic difference between different placing settings.  
For example, it seems unrealistic to assume placing dishes into a rack and
hanging martini glasses upside down on a stemware holder share exactly 
the same hypothesis, although
they might agree on a subset of attributes. While some 
attributes may be shared across different objects and placing areas,
there are some attributes that are specific to the particular 
setting. In such a scenario, it is not sufficient to have either one single model
or several completely independent models for each placing setting
that tend to suffer from over-fitting. Therefore, we
propose to use a shared sparsity structure in our learning.

Say, we have $M$ objects and $N$ placing areas, thus making a total
of $r=MN$ placing `tasks' of particular object-area pairs. Each task can 
have its own model but intuitively these should share some parameters
underneath. To quantify this constraint, we use ideas from  recent works
\citep{jalalidirty,theta_mrf} that attempt to capture the structure in the parameters
of the models. \cite{jalalidirty} used a shared sparsity
structure for multiple linear regressions. We apply their model to the classic
soft-margin SVM. 

In detail, for $r$ tasks, let $\Phi_i\in \Re^{p \times n_i}$ and $Y_i$ denote
training data and its corresponding label, where $p$ is the size of the feature vector 
and $n_i$ is the number of data points in task $i$. We associate every task with a weight vector $\theta_i$. 
We decompose $\theta_i$ in two parts $\theta_i = S_i + B_i$: 
the self-owned features $S_i$ and the shared features $B_i$.
All self-owned features, $S_i$, should have only a few non-zero values so that they can reflect 
individual differences to some extent but would not become dominant in the final model. 
Shared features, $B_i$, need not have identical values, but should share similar sparsity
structure across tasks. In other words, for each feature, they should all either be  
 active or non-active. Let $\left\|S\right\|_{1,1} = \sum_{i,j}|S^j_i|$ and
$\left\|B\right\|_{1,\infty} = \sum_{j=1}^p \max_i |B_i^j|$. Our
new goal function is now:
\begin{eqnarray}
\min_{\theta_i, b_i, i=1,\ldots, r} &  \sum_{i=1}^r \left( \frac{1}{2} \left\|\theta_i\right\|^2_2 + C \sum_{j=1}^{n_i} \xi_{i,j} \right) + \cr
& \qquad \quad \lambda_S \left\|S\right\|_{1,1} + \lambda_B \left\|B\right\|_{1,\infty}\cr
\text{subject to} & Y_i^j(\theta_i^T \Phi_i^j+b_i) \geq 1-\xi_{i,j}, \quad \xi_{i,j}\geq 0\cr
& \qquad \forall 1\leq i\leq r, 1\leq j \leq n_i\cr
& \theta_i = S_i + B_i, \quad \forall 1\leq i\leq r
\label{eq:dirty}
\end{eqnarray}
When testing in a new scenario, different models vote to determine the best placement. 

While this modification results in superior performance with new objects in new placing areas,
it requires one model per object-area pair and therefore it does not scale to a large
number of objects and placing areas.


\begin{table}[t!]
\centering
\caption{{\small Average performance of our algorithm using different features on the SESO scenario.}}
\label{tbl:features}
\begin{tabular}{l c c c c c}
\toprule%
	&	chance	&	contact	&	caging	&	histogram	& all		\\
\midrule
$R_0$	&	29.4	&	1.4	    &	1.9	    &	1.3	    &	\bf{1.0}	\\
P@5	&	0.10	&	0.87	&	0.77	&	0.86	&	\bf{0.95}	\\
AUC 	&   0.54    &	0.89	&	0.83	&	0.86	&	\bf{0.95}	\\
\bottomrule%
\end{tabular}
\end{table}

\section{Experiments on Placing Single Objects}\label{sec:exp_single}
We perform experiments on placing a single object in a designated placing area. 
The dataset includes 8 different objects and 7 placing areas. 
In these experiments, our main purpose is
to analyze the performance of our learning approach in finding
stable placements with preferred orientations. 
Section~\ref{sec:exp_multi} describes our full experiments
with placing multiple objects in complete 3D scenes.

\subsection{Data}
\label{sec:exp_single_data}
Our dataset contains 7 placing areas (3 racks, a flat surface, pen
holder, stemware holder and hook) and 8 objects (mug, martini
glass, plate, bowl, spoon, candy cane, disc and tuning fork).
We generated one training and one test dataset for each object-environment pair.
Each training/test dataset contains 1800 random placements with different locations and orientations. After eliminating placements that have collisions, we have
37655 placements in total. 

These placements were labeled by rigid-body simulation (Fig.~\ref{fig:4_scenes}) and 
then used for our supervised learning algorithm. 
Simulation enabled us to generate massive amounts of labeled data. However, the simulation
itself had no knowledge of placing preferences. When creating the ground-truth
training data, we manually labeled all the stable (as verified by the
simulation) but non-preferred placements as negative examples.

\begin{table*}[htb!]
\caption{\small {\bf Learning experiment statistics:}  The performance of different learning algorithms in different scenarios
is shown. 
The top three rows are the results for baselines, where no training data is used. The fourth row is
trained and tested for the SESO case. The last three rows are trained using joint, independent
and shared sparsity SVMs respectively for the NENO case. }
 \label{tbl:offline_results}
 \begin{center}
{\footnotesize
\newcolumntype{P}[2]{>{\footnotesize#1\hspace{0pt}\arraybackslash}p{#2}}
\setlength{\tabcolsep}{2pt}
\begin{minipage}{1\linewidth}
\resizebox{\hsize}{!} {
\begin{tabular}{cc|ccc| ccc| ccc| ccc| ccc| ccc| ccc| ccc || ccc}
\multicolumn{29}{c}{Listed object-wise, averaged over the placing areas.}\\
\whline{1.1pt}
&& \multicolumn{3}{c| }{plate} & \multicolumn{3}{c| }{mug} & \multicolumn{3}{c| }{martini} &
\multicolumn{3}{c| }{bowl} &   \multicolumn{3}{c| }{candy cane} &\multicolumn{3}{c| }{disc} &
\multicolumn{3}{c| }{spoon}& \multicolumn{3}{c||}{tuning fork} & && \\
\cline{3-26}
 && \multicolumn{3}{c|}{flat,} 
  & \multicolumn{3}{c|}{flat,} 
  & \multicolumn{3}{c|}{flat, 3 racks,} 
  & \multicolumn{3}{c|}{flat,} 
  & \multicolumn{3}{c|}{flat, hook,}
  & \multicolumn{3}{c|}{flat, hook,} 
  & \multicolumn{3}{c|}{flat,}
  & \multicolumn{3}{c||}{flat,} 
  & \multicolumn{3}{c}{\bf{Average}}\\
& & \multicolumn{3}{c|}{3 racks} 
  & \multicolumn{3}{c|}{3 racks} 
  & \multicolumn{3}{c|}{stemware holder}
  & \multicolumn{3}{c|}{3 racks} 
  & \multicolumn{3}{c|}{pen holder}
  & \multicolumn{3}{c|}{pen holder}
  & \multicolumn{3}{c|}{pen holder}
  & \multicolumn{3}{c||}{pen holder}
  &&&
  \\
\whline{0.8pt} 
 &  & $R_0$ & P@5 & AUC & $R_0$ & P@5 & AUC & $R_0$ & P@5 & AUC& $R_0$ & P@5 & AUC& $R_0$ & P@5 & AUC& $R_0$ & P@5 & AUC& $R_0$ & P@5 & AUC& $R_0$ & P@5 & AUC & $R_0$ & P@5 & AUC\\
\whline{0.8pt} 
\multirow{3}{*}{\rotatebox{90}{baseline} }
& chance	& 4.0	& 0.20	& 0.49	& 5.3	& 0.10	& 0.49	& 6.8	& 0.12	& 0.49	& 6.5	& 0.15	& 0.50	& 102.7	& 0.00	& 0.46	& 32.7	& 0.00	& 0.46	& 101.0	& 0.20	& 0.52	& 44.0	& 0.00	& 0.53	& 29.4	& 0.10	& 0.49\\
& flat-up	& 4.3	& 0.45	& 0.38	& 11.8	& 0.50	& 0.78	& 16.0	& 0.32	& 0.69	& 6.0	& 0.40	& 0.79	& 44.0	& 0.33	& 0.51	& 20.0	& 0.40	& 0.81	& 35.0	& 0.50	& 0.30	& 35.5	& 0.40	& 0.66	& 18.6	& 0.41	& 0.63\\
& lowest 	& 27.5	& 0.25	& 0.73	& 3.8	& 0.35	& 0.80	& 39.0	& 0.32	& 0.83	& 7.0	& 0.30	& 0.76	& 51.7	& 0.33	& 0.83	& 122.7	& 0.00	& 0.86	& 2.5	& 0.50	& 0.83	& 5.0	& 0.50	& 0.76	& 32.8	& 0.30	& 0.80\\
\hline
& SESO		& 1.3	& 0.90	& 0.90	& 1.0	& 0.85	& 0.92	& 1.0	& 1.00	& 0.95	& 1.0	& 1.00	& 0.92	& 1.0	& 0.93	& 1.00	& 1.0	& 0.93	& 0.97	& 1.0	& 1.00	& 1.00	& 1.0	& 1.00	& 1.00	& 1.0	& 0.95	& 0.95\\
\hline
\multirow{3}{*}{\rotatebox{90}{NENO} }
& joint		& 8.3	& 0.50	& 0.78	& 2.5	& 0.65	& 0.88	& 5.2	& 0.48	& 0.81	& 2.8	& 0.55	& 0.87	& 16.7	& 0.33	& 0.76	& 20.0	& 0.33	& 0.81	& 23.0	& 0.20	& 0.66	& 2.0	& 0.50	& 0.85	& 8.9	& 0.47	& 0.81\\
& indep.	& 2.0	& 0.70	& 0.86	& 1.3	& 0.80	& 0.89	& 1.2	& 0.86	& 0.91	& 3.0	& 0.55	& 0.82	& 9.3	& 0.60	& 0.87	& 11.7	& 0.53	& 0.88	& 23.5	& 0.40	& 0.82	& 2.5	& 0.40	& 0.71	& 5.4	& 0.64	& 0.86\\
& shared	& 1.8	& 0.70	& 0.84	& 1.8	& 0.80	& 0.85	& 1.6	& 0.76	& 0.90	& 2.0	& 0.75	& 0.91	& 2.7	& 0.67	& 0.88	& 1.3	& 0.73	& 0.97	& 7.0	& 0.40	& 0.92	& 1.0	& 0.40	& 0.84	& 2.1	& 0.69	& 0.89\\
\whline{1.1pt} 
\end{tabular}
}
\end{minipage}
\vskip .1in
\begin{minipage}{0.9\linewidth}
\resizebox{\hsize}{!} {
\begin{tabular}{cc|ccc| ccc| ccc| ccc| ccc| ccc| ccc || ccc}
\multicolumn{26}{c}{Listed placing area-wise, averaged over the objects.}\\
\whline{1.1pt}
&& \multicolumn{3}{c| }{rack1} & \multicolumn{3}{c| }{rack2} & \multicolumn{3}{c| }{rack3} &
\multicolumn{3}{c| }{flat} &   \multicolumn{3}{c| }{pen holder} &\multicolumn{3}{c| }{hook} &
\multicolumn{3}{c|| }{stemware holder}& \multicolumn{3}{c}{}  \\
\cline{3-23}
 && \multicolumn{3}{c|}{plate, mug,} 
  & \multicolumn{3}{c|}{plate, mug,} 
  & \multicolumn{3}{c|}{plate, mug,} 
  & \multicolumn{3}{c|}{all } 
  & \multicolumn{3}{c|}{candy cane, disc,}
  & \multicolumn{3}{c|}{candy cane,} 
  & \multicolumn{3}{c||}{martini}
  & \multicolumn{3}{c}{\bf{Average}}\\
& & \multicolumn{3}{c|}{martini, bowl} 
  & \multicolumn{3}{c|}{martini, bowl} 
  & \multicolumn{3}{c|}{martini, bowl} 
  & \multicolumn{3}{c|}{objects} 
  & \multicolumn{3}{c|}{spoon, tuningfork}
  & \multicolumn{3}{c|}{disc}
  & \multicolumn{3}{c||}{}
  &&&
  \\
\whline{0.8pt} 
 &  & $R_0$ & P@5 & AUC & $R_0$ & P@5 & AUC & $R_0$ & P@5 & AUC& $R_0$ & P@5 & AUC& $R_0$ & P@5 & AUC& $R_0$ & P@5 & AUC& $R_0$ & P@5 & AUC & $R_0$ & P@5 & AUC\\
\whline{0.8pt} 
\multirow{3}{*}{\rotatebox{90}{baseline} }
& chance	& 3.8	& 0.15	& 0.53	& 5.0	& 0.25	& 0.49	& 4.8	& 0.15	& 0.48	& 6.6	& 0.08	& 0.48	& 128.0	& 0.00	& 0.50	& 78.0	& 0.00	& 0.46	& 18.0	& 0.00	& 0.42	& 29.4	& 0.10	& 0.49\\
& flat-up   & 2.8	& 0.50	& 0.67	& 18.3	& 0.05	& 0.47	& 4.8	& 0.20	& 0.60	& 1.0	& 0.98	& 0.91	& 61.3	& 0.05	& 0.45	& 42.0	& 0.00	& 0.45	& 65.0	& 0.00	& 0.58		& 18.6	& 0.41	& 0.63\\
& lowest    & 1.3	& 0.75	& 0.87	& 22.0	& 0.10	& 0.70	& 22.0	& 0.15	& 0.80	& 4.3	& 0.23	& 0.90	& 60.3	& 0.60	& 0.85	& 136.5	& 0.00	& 0.71	& 157.0	& 0.00	& 0.56		& 32.8	& 0.30	& 0.80\\
\hline
& SESO      & 1.0	& 1.00	& 0.91	& 1.3	& 0.75	& 0.83	& 1.0	& 1.00	& 0.93	& 1.0	& 0.95	& 1.00	& 1.0	& 1.00	& 0.98	& 1.0	& 1.00	& 1.00	& 1.0	& 1.00	& 1.00		& 1.0	& 0.95	& 0.95\\
\hline
\multirow{3}{*}{\rotatebox{90}{NENO} }
& joint     & 1.8	& 0.60	& 0.92	& 2.5	& 0.70	& 0.84	& 8.8	& 0.35	& 0.70	& 2.4	& 0.63	& 0.86	& 20.5	& 0.25	& 0.76	& 34.5	& 0.00	& 0.69	& 18.0	& 0.00	& 0.75& 8.9	& 0.47	& 0.81\\
& indep.    & 1.3	& 0.70	& 0.85	& 2.0	& 0.55	& 0.88	& 2.5	& 0.70	& 0.86	& 1.9	& 0.75	& 0.89	& 12.3	& 0.60	& 0.79	& 29.0	& 0.00	& 0.79	& 1.0	& 1.00	& 0.94		& 5.4	& 0.64	& 0.86\\
& shared    & 1.8	& 0.75	& 0.88	& 2.0	& 0.70	& 0.86	& 2.3	& 0.70	& 0.84	& 1.3	& 0.75	& 0.92	& 4.0	& 0.60	& 0.88	& 3.5	& 0.40	& 0.92	& 1.0	& 0.80	& 0.95	& 2.1	& 0.69	& 0.89\\
\whline{1.1pt} 
\end{tabular}
}
\end{minipage}
}
\end{center}
\end{table*}

\subsection{Learning Scenarios\label{sec:learning_scenario}}
In real-world placing, the robot may or may not encounter new placing areas and new objects. 
Therefore, we trained our algorithm for two 
different scenarios: 1) Same Environment Same Object (\textbf{SESO}), where training data only
contains the object and the placing environment to be tested. 
2) New Environment New Object (\textbf{NENO}). In this case, the training data includes all other
objects and environments except the one for test. 
We also considered two additional learning scenarios, SENO and NESO, in \cite{JiangPlacing}.
More results can be found there.

\subsection{Baseline Methods}
We compare our algorithm with the following three heuristic methods:

\begin{itemize}

\item
\textit{Chance.} The location and orientation is randomly sampled within the bounding box of
the area and guaranteed to be `collision-free.'

\item\textit{Flat-surface-upright rule.}
Several methods exist for detecting `flat' surfaces \citep{kemp-placing}, and we consider a 
placing method based on finding flat surfaces.
In this method, objects would be placed with \textit{pre-defined} upright orientation on the surface.
When no flat surface can be found, a random placement would be picked. Note that this heuristic
actually uses \textit{more} information than our method.

\item
\textit{Finding lowest placing point.} 
For many placing areas, such as dish-racks or containers, a lower placing point often gives more
stability. Therefore, this heuristic rule chooses the placing point with the lowest height.

\end{itemize}

\subsection{Evaluation Metrics}\label{sec:eval_single}
We evaluate our algorithm's performance on the following metrics:
\begin{itemize}
\item $R_0$: Rank of the first valid placement. 
($R_0=1$ ideally)
\item P@5: In top 5 candidates, the fraction of valid placements. 
\item AUC: Area under ROC Curve \citep{hanley_AUC}, a classification metric computed from all the candidates.
\item $P_\text{stability}$: Success-rate (in \%) of placing the 
object stably with the robotic arm.
\item $P_\text{preference}$: Success-rate (in \%) of placing the object stably in preferred
configuration with the robotic arm.
\end{itemize}

\subsection{Learning Experiments\label{sec:offline_experiments}}

We first verified that having different types of features is helpful in performance,
as shown in Table~\ref{tbl:features}.
While all three types of features outperform chance, combining them together gives
the best results under all evaluation metrics.  

Next, Table~\ref{tbl:offline_results} shows the comparison of three
heuristic methods and three variations of SVM learning algorithms: 1) joint SVM, where one single
model is learned from all the placing tasks in the training dataset; 2) independent SVM, which
treats each task as a independent learning problem and learns a separate model per task; 3) shared
sparsity SVM (Section~\ref{sec:shared_sparsity_svm}), which also learns one model per task but with parameter sharing. 
Both independent and
shared sparsity SVM  use voting to rank placements for the test case.

Table~\ref{tbl:offline_results} shows
that all the learning methods (last four rows) 
outperform heuristic rules under all evaluation metrics.
Not surprisingly, the chance method performs poorly (with Prec@5=$0.1$ and AUC=$0.49$) because 
there are very few preferred placements in the large sampling space of possible placements.
The two heuristic methods perform well in some obvious cases such as using flat-surface-upright
method for table or lowest-point method for rack1. 
However,  their performance varies significantly in non-trivial cases,
including the stemware holder and the hook.
This demonstrates that it is hard to script a placing rule that works universally.

We get close to perfect results for the SESO case---i.e., the learning algorithm
can very reliably predict object placements if a known object was being 
placed in a previously seen location.   
The learning scenario NENO is extremely challenging---here, for each task (of an 
object-area pair), the algorithm is trained without either the object
or the placing area in the training set. In this case, $R_0$ increases from $1.0$ to $8.9$ with joint
SVM, and to $5.4$ using independent SVM with voting.
However, shared sparsity SVM 
(the last row in the table) helps to reduce the average $R_0$ down to $2.1$.
While shared sparsity SVM outperforms other algorithms, 
the result also indicates that independent SVM with voting is better than
joint SVM. This could be due to the large variety in the placing situations
in the training set. Thus imposing one model for all tasks decreases the performance. 
We also observed that in cases where the placing strategy is very different from the ones trained on, 
the shared sparsity SVM does not perform well. For example, $R_0$ is $7.0$ for 
spoon-in-pen-holder and is $5.0$ for disk-on-hook.
This issue could potentially be  addressed by expanding the training dataset.
For comparison, the average AUC (area under ROC curve) of shared sparsity SVM is $0.89$, which
compares to $0.94$ in Table~\ref{tbl:stability} for corresponding classes (dish-racks,
stemware holder and pen holder).

\begin{figure}[t!]
\centering
\includegraphics[height=0.083\textheight]{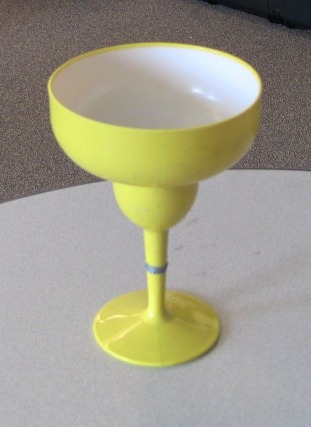}
\includegraphics[height=0.083\textheight]{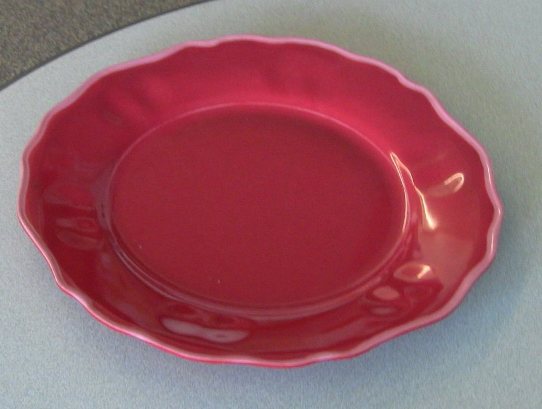}
\includegraphics[height=0.083\textheight]{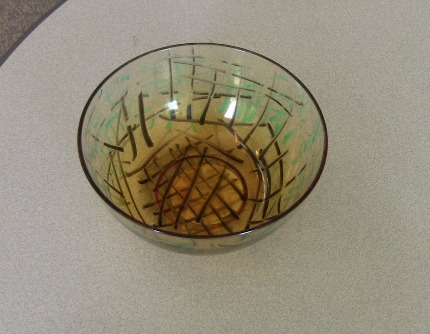}\\
\includegraphics[height=0.083\textheight]{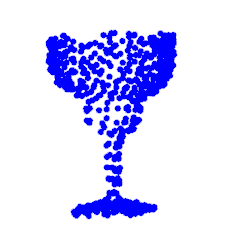}
\includegraphics[height=0.083\textheight]{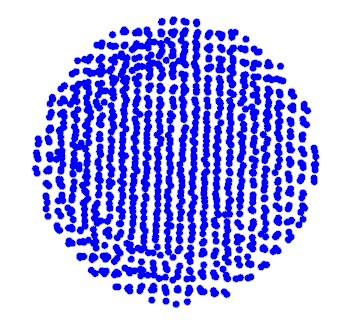}
\includegraphics[height=0.083\textheight]{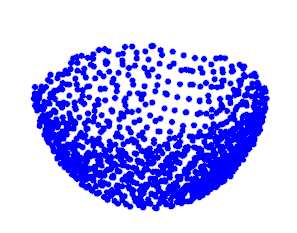}\\
\includegraphics[height=0.083\textheight]{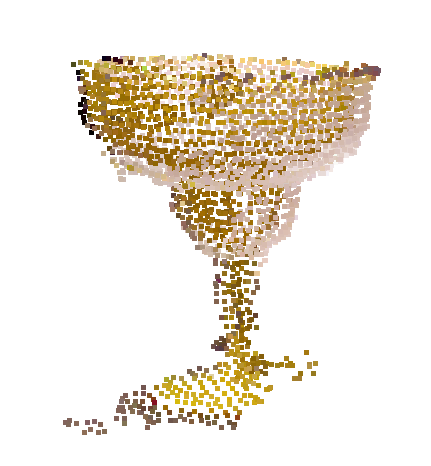}
\includegraphics[height=0.083\textheight]{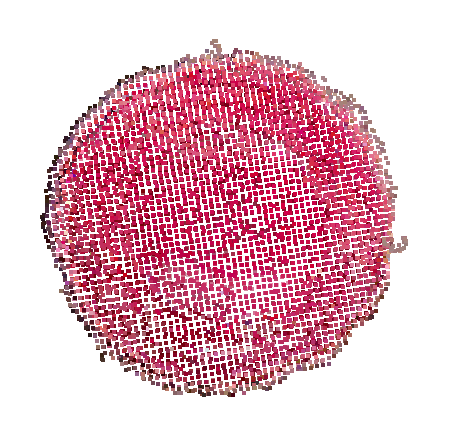}
\includegraphics[height=0.083\textheight]{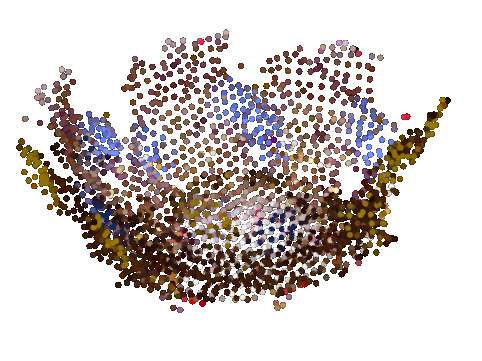}
\caption{{\small Three objects used in our robotic experiments. The top row shows the real objects.
Center row shows the perfect point-clouds extracted from object models. Bottom row shows the raw
point-clouds perceived from the Kinect sensor, used in the robotic experiments.}}
\label{fig:exp_single_robotic_obj}
\end{figure}

\subsection{Robotic Experiments}
\label{sec:exp_single_robotic}

\begin{table*}[t!]
\centering
\caption{{\small Robotic experiments. The algorithm is trained
using shared sparsity SVM under the two learning scenarios: SESO and NENO. 10 trials each are
performed for each object-placing area pair. $P_s$ stands for $P_\text{stability}$ and $P_p$ stands for $P_\text{preference}$.
In the experiments with object models, $R_0$ stands for
the rank of first predicted placements passed the stability test. In the experiments without object
models, we do not perform stability test and thus $R_0$ is not applicable. 
In summary, robotic experiments show a success rate of 98\% when the object has been
seen before and its 3D model is available, and show a success-rate of 82\% (72\% when also
considering semantically correct orientations) when the object has not been seen by the robot before in any form.
}} 
\label{tbl:online_results}
{\footnotesize
\begin{tabular}{c|c|l| ccc |cccc| ccc| c}
\whline{1.1pt}
\multicolumn{3}{c|}{}& \multicolumn{3}{c|}{plate} & \multicolumn{4}{c|}{martini} & \multicolumn{3}{c|}{bowl} &\multirow{2}{*}{\bf{Average}}\\
\multicolumn{3}{c|}{}& rack1 & rack3 & flat& rack1 & rack3 & flat& stem. & rack1 & rack3 & flat&\\
\whline{0.8pt}
\multirow{6}{*}{\rotatebox{90}{\textbf{w/} obj models}} &
\multirow{3}{*}{{SESO}} &
$R_0$   & 1.0	& 1.0	& 1.0	& 1.0	& 1.0	& 1.0	& 1.0	& 1.0	& 1.0	& 1.0	& \bf{1.0}\\
&&
$P_s$(\%)   & 100	& 100	& 100	& 100	& 100	& 100	& 80	& 100	& 100	& 100	& \bf{98}\\
&&
$P_p$(\%)   & 100	& 100	& 100	& 100	& 100	& 100	& 80	& 100	& 100	& 100	& \bf{98}\\
\cline{2-14}
&\multirow{3}{*}{{NENO}} &
$R_0$   & 1.8	& 2.2	& 1.0	& 2.4	& 1.4	& 1.0	& 1.2	& 3.4	& 3.0	& 1.8	& 1.9\\
&&
$P_s$(\%)   & 100	& 100	& 100	& 80	& 100	& 80	& 100	& 100	& 100	& 100	& 96\\
&&
$P_p$(\%)   & 100	& 100	& 100	& 60	& 100	& 80	& 100	& 80	& 100	& 100	& 92\\
\whline{0.6pt}
\multirow{6}{*}{\rotatebox{90}{\textbf{w/o} obj models}} &
\multirow{3}{*}{{SESO}} &
$R_0$   & -		& -		& -		& -		& -		& -		& -		& -		& -		& -		& -\\
&&
$P_s$(\%)   & 100	& 80	& 100	& 80	& 100	& 100	& 80	& 100	& 100	& 100	& 94\\
&&
$P_p$(\%)   & 100	& 80	& 100	& 80	& 80	& 100	& 80	& 100	& 100	& 100	& 92\\
\cline{2-14}
&\multirow{3}{*}{{NENO}} &
$R_0$   & -		& -		& -		& -		& -		& -		& -		& -		& -		& -		& -\\
&&
$P_s$(\%)   & 80	& 60	& 100	& 80	& 80	& 100	& 70	& 70	& 100	& 80	& \bf{82}\\
&&
$P_p$(\%)   & 80	& 60	& 100	& 60	& 70	& 80	& 60	& 50	& 80	& 80	& \bf{72}\\
\whline{1.1pt}
\end{tabular}
}    
\end{table*}

\begin{figure}[t!]
\centering
\includegraphics[height=0.11\textheight]{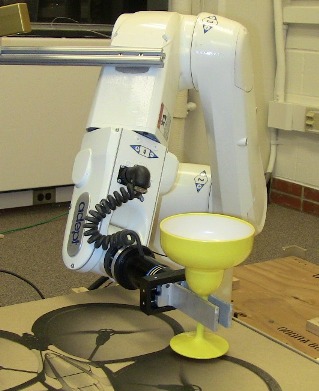}
\includegraphics[angle=90, height=0.11\textheight]{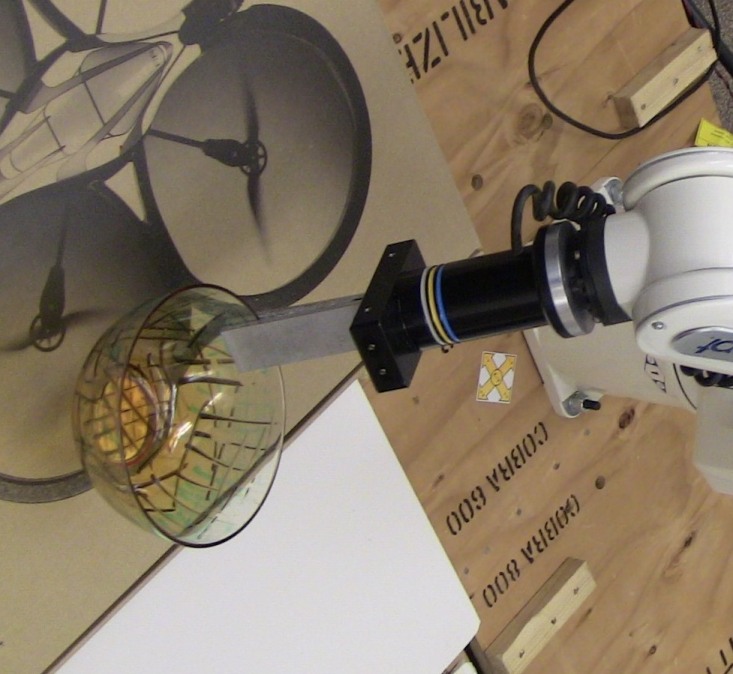}
\includegraphics[height=0.11\textheight,width=0.45\linewidth]{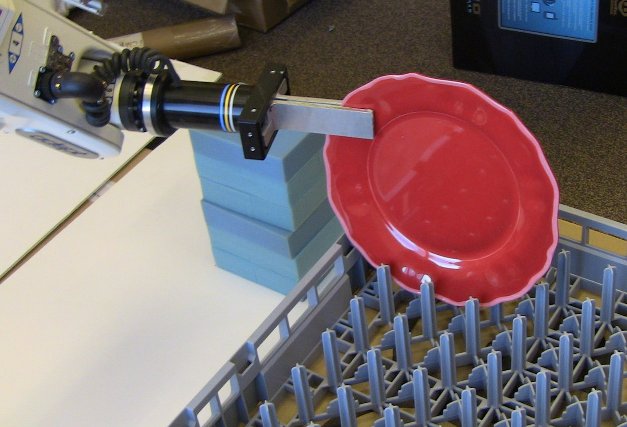}\\
\vspace{.06in}                        
\includegraphics[angle=90, height=0.11\textheight]{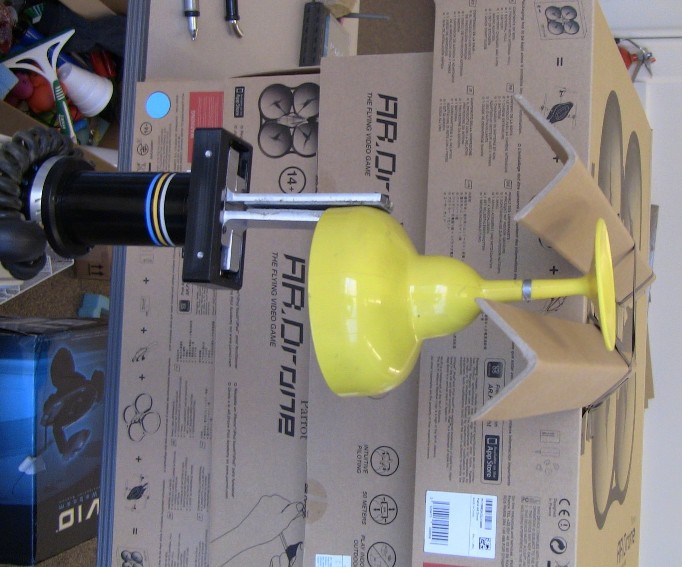}
\includegraphics[height=0.11\textheight]{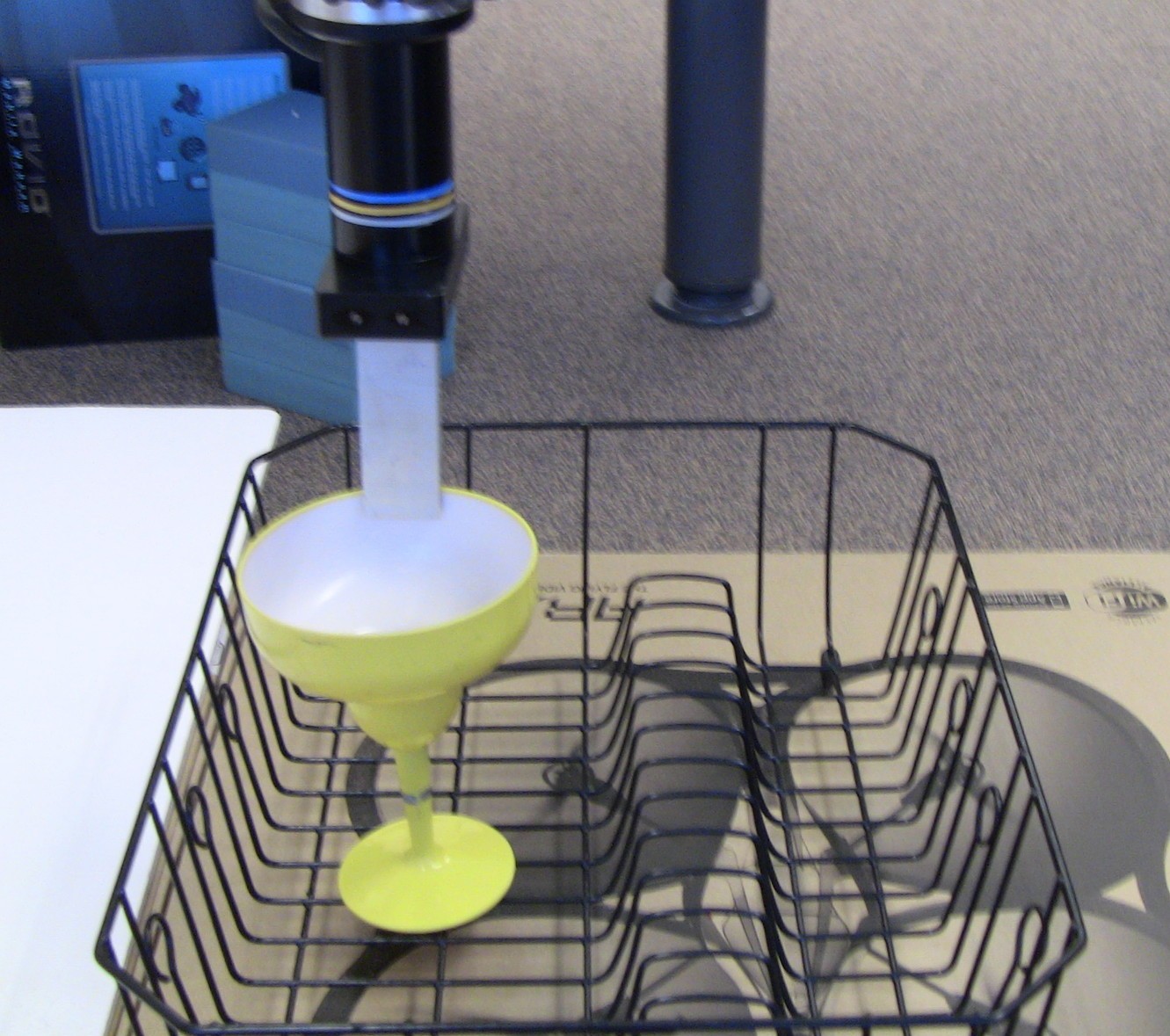}
\includegraphics[height=0.11\textheight]{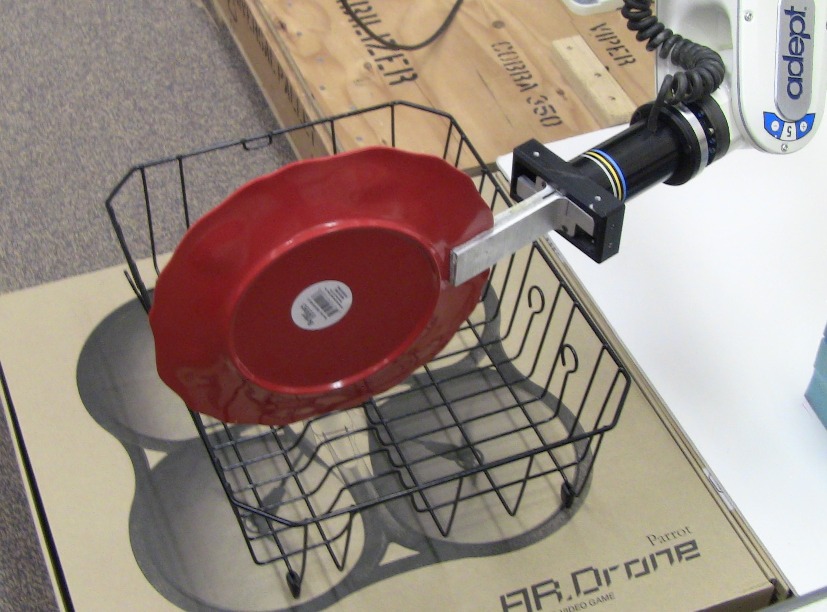}\\
\caption{{\small Robotic arm placing different objects in several placing areas: a martini glass on
a flat surface, a bowl on a flat surface, a plate in rack1, a martini glass on a stemware
holder, a martini glass in rack3 and a plate in rack3. 
}}
\label{fig:screenshots}
\end{figure}

We conducted single-object placing experiments on our PANDA robot with the Kinect sensor,
using the same dataset (Section~\ref{sec:exp_single_data}) for training.
 We tested 10 different placing tasks with 10 trials for each.


In each trial, the robot had to pick up the object and place it in the designated area. 
The input to our algorithm in these experiments was raw point-clouds of the object
and the placing area (see Fig.~\ref{fig:exp_single_robotic_obj}).
Given a placement predicted by our learning algorithm and a feasible grasp, the robot used path planning to
move the object to the destination and released it. A placement was considered
successful if it was stable (the object remained still for more than a minute) and
in its preferred orientation (within $\pm15^\circ$ of the ground-truth
orientation after placing).


Table~\ref{tbl:online_results} shows the results for three objects being
placed by the robotic arm in four placing scenarios (see the bottom six rows).  
We obtain a 94\% success rate in placing the objects stably in SESO case,
and 92\% if we disqualify those stable placements that were not preferred ones.
In the NENO case, we achieve 82\% performance for stable
placing, and 72\% performance for preferred placing. 
Figure~\ref{fig:screenshots} shows several screenshots of our robot placing the objects. 
There were some cases where
 the martini glass and the bowl were placed
horizontally in rack1. 
In these cases, even though the placements were stable, they were not counted as preferred. 
Even small displacement errors while inserting the martini glass in the narrow slot of
the stemware holder often resulted in a failure. In general, several failures for the
bowl and the martini glass were due to incomplete capture of the point-cloud which
resulted in the object hitting the placing area (e.g., the spikes in the dish-rack). 

%

In order to analyze the source of the errors in robotic placing, we did another
experiment in which we factored away the errors caused by the incomplete point-clouds.
In detail, we recovered the full 3D geometry by registering the raw point-cloud
against a few parameterized objects in a database using the Iterative 
Closest Point (ICP) algorithm \citep{ICP1,ICP2}.  Furthermore, since we had
 access to a full solid 3D model, we verified the stability of the placement 
using the rigid-body simulation before executing it on the robot.
If it failed the stability test, we would try the placement with the next highest score
until it would pass the test.
Even though this simulation was computationally 
expensive,\footnote{A single stability test takes $3.8$ second on a $2.93$GHz dual-core processor on average.}
we only needed to compute this for a few top scored placements.

In this setting with a known library of object models, we obtain a 98\% success rate in
placing the objects in SESO case. The robot failed only in one experiment, when the martini glass could not fit into the
narrow stemware holder due to a small displacement occurred in grasping. 
In the NENO case, we achieve 96\% performance in stable placements
and 92\% performance in preferred placements. 
This experiment indicates that with better sensing of the point-clouds, 
our learning algorithm
can give better performance.

\section{Algorithm for Placing Multiple Objects}
\label{sec:alg_multi}
We will now describe our approach for placing multiple objects in a scene,
where we also need to decide which placing area in the scene is semantically
suitable for placing every object in. 


\begin{figure}
\centering
\includegraphics[width=1.0\linewidth]{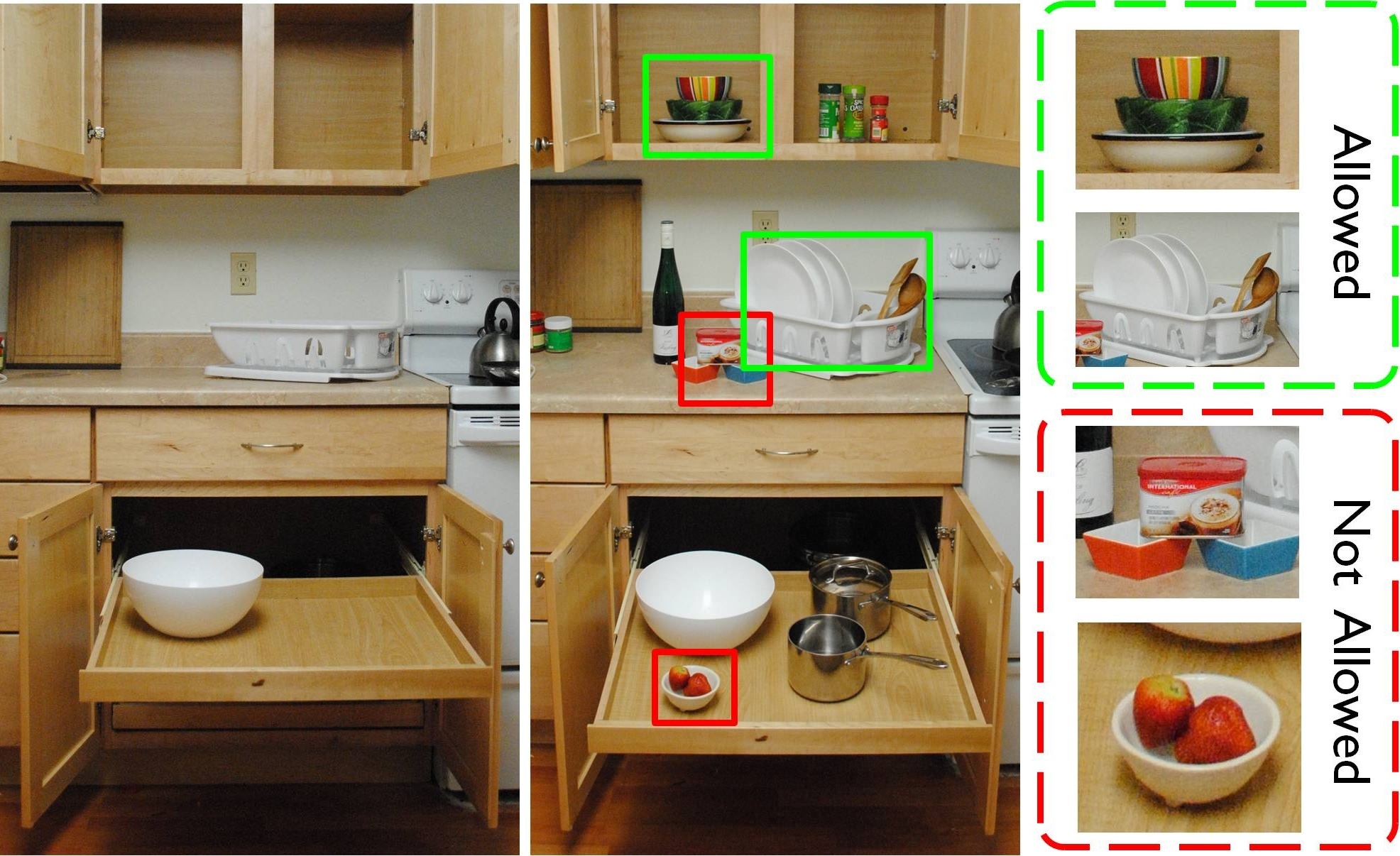}
\caption{{\small Given an initial kitchen scene (left), a possible placing
strategy for multiple objects could be as shown in the middle image: loading the dish-rack with spatulas and
plates, or stacking them up in the cabinet, storing saucepans on the bottom drawer,
etc. In this paper, we only allow chain stacking (see text in Section~\ref{sec:alg_multi}), 
which allows most but not all the possible placing situations (right column).}}

\label{fig:kitchen_example}
\end{figure}

We first state our assumptions in this setting. 
We consider two scenarios: 1) the object is placed
directly on the placing area; 2) the object is stacked on another object. 
While an unlimited number of objects can stack into one 
pile in series (e.g., the plates and bowls in the cabinet in Fig.~\ref{fig:kitchen_example}), 
we do not allow more than one object to be placed on one single object
and we do not allow one object to be placed on more than one object. We refer this assumption as
\textbf{`chain stacking'}. 
For instance, in Fig.~\ref{fig:kitchen_example}, it is not allowed to place two strawberries 
in the small bowl, 
nor is it allowed to place a box on top of the blue and the orange sauce bowls at the same time.
Note that this constraint does not limit placing
multiple objects on a placing area directly, e.g., the dish-rack is given
as a placing area and therefore we can place multiple plates in it.


A physically feasible placing strategy needs to satisfy two constraints:
1) \textbf{Full-coverage}: every given object must be placed somewhere, either directly on a placing
area or on top of another object.
2) \textbf{Non-overlap}: two objects cannot be placed at same location, although being in the same
placing area is allowed. For example, multiple plates can be loaded into one
dish-rack, however, they cannot occupy the same slot at the same location. 
3) \textbf{Acyclic}: placing strategy should be acyclic, i.e., if object A is on top of B (either
directly or indirectly), then B cannot be above A.

\begin{figure}[tb!]
\centering
\includegraphics[width=0.8\linewidth]{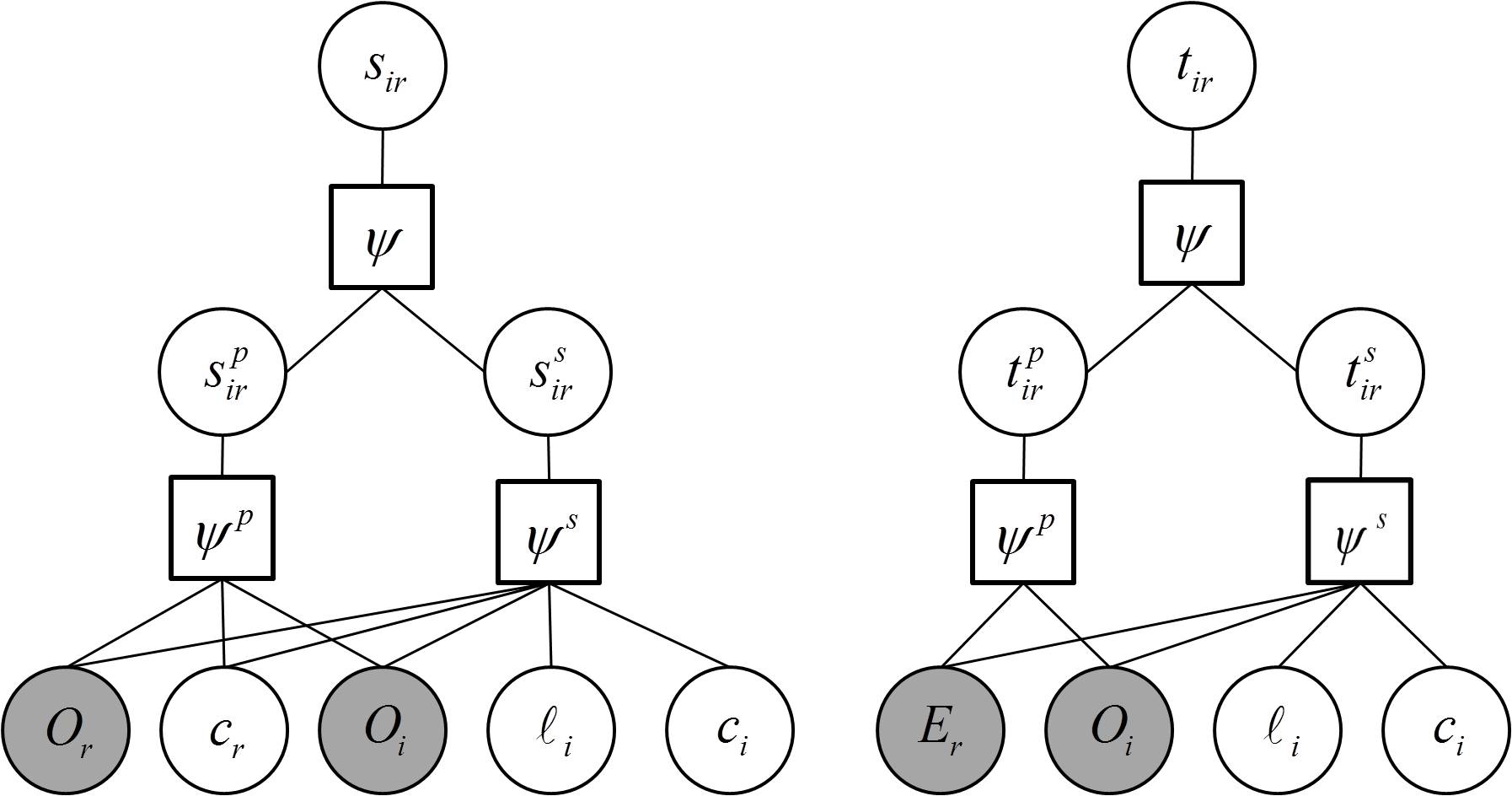}
\caption{\small{ Graphical models for two types of single placements: stacking on another object (left) and directly
    placing on an environment (right). The shaded nodes are observed point-clouds. }}
\label{fig:gadget}
\end{figure}

\subsection{Graphical Model}

We now introduce our representation of a placing strategy as a graphical model, followed by our
design of features, max-margin learning algorithm, and inference as a linear programming problem. 

As we mentioned in Section~\ref{sec:def_multi}, given $n$ objects $\mathcal
O=\{O_1, \ldots,O_n\}$ and $m$ placing areas (or environments) $\mathcal E=\{E_1,\ldots,E_m\}$
our goal is to find a placing strategy, that can be specified by a 
tuple $(S, T, C, L)$: 
\begin{itemize}
\item $S=\{s_{ir}\in\{0,1\}|1\leq i\leq n, 1\leq r\leq n \}$: whether $O_i$ is
stacking on top of $O_r$.
\item $T=\{t_{ir}\in\{0,1\}|1\leq i\leq n, 1\leq r\leq m \}$: whether $O_i$ is
put directly on $E_r$.
\item $C=\{c_i\in \text{SO(3)} | 1\leq i \leq n\}$: the configuration (i.e., 3D orientation) of $O_i$.
\item $L=\{\ell_i\in \Re^3 | 1\leq i \leq n\}$: the 3D location of $O_i$ w.r.t. its base. 
\end{itemize}

We now design a potential function over all placing strategies given a scene, i.e., $\Psi(S,
T, L, C , \mathcal O, \mathcal E)$. This function reflects the placing quality
and we associate higher potential to better placing strategies. 
The best strategy is the one with the maximum potential. Therefore, our goal is,
\begin{equation}\label{eq:goal}
      (S^*, T^*, L^*, C^*)=\arg\max_{S,T,L,C}\Psi(S,T,L,C,\mathcal O,\mathcal E) 
\end{equation}     
where the solution $(S,T,L,C)$ should follow certain constraints that we will discuss in
Section~\ref{sec:alg_multi_inference}. 


We use an undirected graphical model (Markov networks~\citep{koller2009probabilistic}) for defining the potential.
Fig.~\ref{fig:gadget} shows two simplified graphical models for a single
placement. The entire graphical model for multiple objects and placing areas 
is an assembly of these basic structures (one for each $s_{ir}$ and
$t_{ir}$) with shared $O,E,\ell$ and $c$ nodes.
We now explain this graphical model and show how to factorize $\Psi(S,T, L, C ,
\mathcal O, \mathcal E)$ into small pieces so that learning and inference is
tractable.
 
Our graphical model indicates the following independence: 
\begin{itemize}
\item  $s_{ir}$ and $t_{ir}$ are conditionally independent given $L, C, \mathcal O,
\mathcal E$.

\item $s_{ir}$($t_{ir}$) only depends on objects (environments) involved in this
single placing task of placing $O_i$ on $O_r$ ($E_r$). This implies that, 
when placing an object $i$ on object $j$, it does not matter where object $j$
is placed.
\end{itemize}
Consequently, we can factorize the overall potential as,
\begin{eqnarray}
 \Psi(S, T, L, C, \mathcal O, \mathcal E) & = & \prod_{i,r} \Psi(s_{ir},O_i, \ell_i,
         c_i, O_r, c_r)\notag\\
    &&\times\prod_{i,r} \Psi(t_{ir}, O_i, \ell_i, c_i, E_r) \label{eq:indep}
\end{eqnarray}     

We further factorize the potential of each placement into three terms to encode
the stability and semantic preference in placing, as shown in Fig.~\ref{fig:gadget}.
For each placement, we introduce two binary variables indicating its stability (with
superscript $s$) and semantic preference (with superscript $p$). Considering
they are latent, we have,
\begin{align}
&\scriptstyle \Psi(s_{ir}, O_i, \ell_i, c_i, O_r,c_r) = \nonumber\\
&\scriptstyle \sum_{s^s_{ir}, s^p_{ir}\in\{0,1\}} \psi(s_{ir} , s^s_{ir},
		s^p_{ir})\psi^s(s^s_{ir},O_i, \ell_i, c_i, O_r, c_r)\psi^p(s^p_{ir},O_i, O_r,
            c_r)\nonumber\\
&\scriptstyle \Psi(t_{ir},O_i, \ell_i, c_i, E_r) =\nonumber\\
&\scriptstyle \sum_{t^s_{ir},t^p_{ir}\in\{0,1\}} \psi(t_{ir} , t^s_{ir},
		t^p_{ir})\psi^s(t^s_{ir},O_i, \ell_i, c_i, E_r)\psi^p(s^p_{ir},O_i,
        E_r)\label{eq:factor}
\end{align} 

The intuition is that a placement is determined by two factors: stability and
semantic preference. This is quantified by the potential function $\psi$. 
$\psi^s$ captures stability which is only dependent on local geometric information, i.e., exact poses and
locations. On the other hand, semantic preference concerns how well this object fits the 
environment. So the function $\psi^p$ is determined by the object and the base regardless of the details of the placement.
For example, placing a plate in a dish-rack has a high semantic preference over the
ground, but different placements (vertical vs. horizontal) would only change its
stability.  
Note that in stacking, semantic preference ($s_{ir}^p$) is also dependent on the
configuration of the base ($c_r$ ), since the base's configuration would affect the context. For
instance, we can place different objects on a book lying horizontally, but
it is hard to place any object on a book standing vertically.

Based on the fact that a good placement should be stable as well as semantically
correct, we set 
$\psi(s_{ir},s^s_{ir}, s^p_{ir}) = 1$ if $s_{ir}=(s^s_{ir} \land s^p_{ir})$
otherwise 0.  We do the same for $\psi(t_{ir},t^s_{ir}, t^p_{ir})$.
As for the potential functions $\psi^s$ and $\psi^p$, they are based on a collection of features
that indicate good stability and semantic preference. Their parameters 
are learned using the max-margin algorithm described in Section~\ref{sec:svm_multi}.

\begin{table}
{\scriptsize
\begin{center}	
\caption{{\small Features for multiple objects placements and their dimensions (`Dim').}}
\label{tbl:feature}
\begin{tabular}{l@{}r l@{}r}
\toprule
Feature descriptions & Dim  &Feature descriptions & Dim  \\
\midrule          
\textbf{Stability} & \textbf{178} & \textbf{Semantic preference} & \textbf{801}\\
Supporting contacts & 12 & Zernike 	& $37\times4$   \\
Caging features &	4	 &BOW	        & $100\times4$  \\
Histograms & 162 & Color histogram       & $46\times4$ 	\\
	&	  & Curvature histogram       & $12\times4$   \\
 	&	  & Overall shape	        & $5\times4$   	\\
 	&	  & Relative height       & 1 	\\
\bottomrule			 
\end{tabular}
\end{center}
}
\end{table}





\subsection{Features}

In the multiple-object placements, we introduce additional semantic features for
choosing a good placing area for an object.
Our stability features $\phi_s$ are similar to those described in 
Section~\ref{sec:feature_single}.\footnote{In multiple-object placement experiments,
we only use the relative height caging features with $n_r=1$ and 
$n_\theta=4$, and use 81-bin ($9\times 9$) grid without the ratios for the histogram features.}
Semantic features $\phi_p$ depend only on $\mathsf O$ and $\mathsf B$,
where $\mathsf O$ denotes the point-cloud of object
$O$ to be placed and $\mathsf B$ denote the point-cloud of the base (either an
environment or another object).  This is because
the semantic features should be invariant to different placements of the object within the same
placing area.  We describe them in the following:


\begin{itemize}
\item \textbf{3D Zernike Descriptors:}
Often the semantic information of an object is encoded in its shape. 
Because of their rotation- and translation-invariant property, we
apply 3D Zernike descriptors~\citep{novotni20033d}
 to $\mathsf O$ and $\mathsf B$ for capturing their geometric information.
 This gives us $37$ values for a point-cloud.

\item \textbf{Bag-of-words (BOW) Features:}
Fast Point Feature Histograms (FPFH) \citep{Rusu09ICRA}
are persistent under different views and point densities and produce different signatures for points
on different geometric surfaces such as planes, cylinders, spheres, etc. We
compute a vocabulary of FPFH signatures by extracting 100 cluster centers from
FPFH of every point in our training set. 
Given a test point-cloud, we compute the FPFH signature for each point and associate it with its
nearest cluster center.  The histogram over these cluster
centers makes our BOW features. 

\item \textbf{Color Histograms:}
The features described above  capture only the geometric cues, but not the visual
information such as color. Color information can give clues to the texture of an
object and can help identify some semantic information.  We 
compute a 2D histogram of hue and saturation ($6\times 6$) and 
a 1D $10$-bin histogram of intensity, thus giving a total of $46$
features.

\item \textbf{Curvature Histograms:}
We estimated the curvature of every point using Point Cloud Library~\citep{Rusu_ICRA2011_PCL}, and then compute
its 12-bin histogram.
 
\item \textbf{Overall Shape:} 
Given a point-cloud, we compute three Eigenvalues of its covariance matrix
($\lambda_1 \geq \lambda_2 \geq \lambda_3$). We then compute their ratios ($\lambda_2/\lambda_1$,
$\lambda_3/\lambda_2$), thus giving us a total of $5$ features.

\end{itemize}

We extract the aforementioned semantic features for both $\mathsf O$ and $\mathsf B$ 
separately. For each, we also add their pairwise product and pairwise minimum,
thus giving us $4$ values for each feature. 
The last feature is the height to the ground (only
for placing areas), thus giving a total of $801$ features.
We summarize the semantic features
in Table~\ref{tbl:feature}.

\subsection{Max-margin Learning}\label{sec:svm_multi}

Following \cite{M3Nets}, we choose the log-linear model for potential functions $\psi^s$ and
$\psi^p$,\footnote{For conciseness, we use $\psi(s^s_{ir})$ to denote the full term
$\psi^s(s^s_{ir}, O_i, \ell_i, c_i, O_r, c_r)$ 
 explicitly in the rest of the paper unless otherwise
clarified. We do the same for $\psi^p$.}
\begin{equation}\log \psi^s(s^s_{ir}) \propto \theta_s^\top \phi_s(O_i, \ell_i, c_i, O_r, c_r)
\end{equation}  %
where $\theta_s$ is the weight vector of the features and 
$\phi_s(\cdot)$ is the feature vector for
stability. Let $\theta_p$ and $\phi_p(\cdot)$ denote the weight and features for semantic preference respectively. 
The discriminant function for a placing strategy is given by (from Eq.~\eqref{eq:indep})
\begin{equation}
f (S,T) = \sum_{ir} \log \Psi(s_{ir})+ \sum_{ir}\log \Psi(t_{ir})
\end{equation}

In the learning phase, we use supervised learning to estimate 
$\theta_s$ and $\theta_p$. The training data contains placements with
ground-truth stability and semantic preference labels (we describe it later in
Section~\ref{sec:exp_multi_data}). By Eq.~\eqref{eq:factor},
\begin{equation}
f (S,T) = \sum_{ir} \theta_s^T \phi_s(s^s_{ir}) + \theta_p^T \phi_p(s^p_{ir})+ 
\sum_{ir} \theta_s^T \phi_s(t^s_{ir}) + \theta_p^T \phi_p(t^p_{ir})
\end{equation}

A desirable $\theta_s$ and $\theta_p$ should
be able to maximize the potential of a good placing strategy $(S,T)$, i.e., 
for any different placing strategy $(S',T')$, $f(S,T) >
f(S',T')$. 
Furthermore, we want to maximize this difference to increase the confidence in the learned model.
Therefore, our objective is,
\begin{equation} \arg\max_{\theta_s, \theta_p} \gamma 
~~~~\text{s.t. }\; \scriptstyle f(S,T) - f(S',T') \geq \gamma, \quad \forall (S',T')\neq (S,T) 
\end{equation}
which, after introducing slack variables to allow errors in the training
data, is equivalent to\footnote{Here, without loss of the generality, we map the domain of $S$ and $T$
from \{0,1\} to \{-1,1\}.}
\begin{align}
&\scriptstyle \arg\min_{\theta_s, \theta_p} \frac{1}{2} (||\theta_s||^2 + ||\theta_p||^2) + C\sum (\xi^s_{s,ir} + \xi^p_{s,ir} + \xi^s_{t,ir} + \xi^p_{t,ir})\\
&\scriptstyle\text{s.t. } s^s_{ir}(\theta_s^\top \phi_s (\cdot))\geq 1-\xi^s_{s,ir},\quad 
    s^p_{ir}(\theta_p^\top \phi_p (\cdot))\geq 1 - \xi^p_{s,ir},  \\
&\scriptstyle \quad t^s_{ir}(\theta_s^\top \phi_s (\cdot))\geq 1-\xi^s_{t,ir}, \quad 
    t^p_{ir}(\theta_p^\top \phi_p (\cdot))\geq 1 - \xi^p_{t,ir}, \forall i,r 
\end{align}    
We use max-margin learning \citep{SVMlight} to learn $\theta_s$ and
$\theta_p$ respectively.  

Note that the learning method in Section~\ref{sec:svm_single} for single-object placements
is actually a special case of the above. Specifically, for one object and one placing area, 
the stacking and the semantic preference problems are trivially solved, and the subscripts
$i$ and $r$ are not needed. Therefore, this equation reduces to Eq.~\eqref{eq:svm_soft}
with $\theta_s$, $\phi_s(\cdot)$ and $t^s$  corresponding to $\theta$, $\phi_i$ and
$y_i$ in Eq.~\eqref{eq:svm_soft}.

\subsection{Inference}\label{sec:alg_multi_inference}
Once we have learned the parameters in the graphical model, given the objects and placing
environment, we need to find the placing strategy that maximizes the likelihood in
Eq.~\eqref{eq:goal} while satisfying the aforementioned constraints as well: 
\begin{eqnarray} \label{eq:constraints1}
\sum_{i=1}^n s_{ir} \leq 1, \quad \forall r;&&\text{(chain stacking)}\quad\\
\sum_{r=1}^n s_{ir}+\sum_{r=1}^m t_{ir} = 1,\quad \forall i;&&\text{(full-coverage)}\label{eq:constraints2}
\\
\ell_i \neq \ell_j, \quad\forall t_{ir}=t_{jr}=1, &&\label{eq:constraints3}
\\
t_{ir}+t_{jr}\leq 1,~\quad \forall \scriptstyle O_i \text{ overlaps } O_j.\quad &&\text{(non-overlap)}
\label{eq:constraints4}
\end{eqnarray}

Eq.~\eqref{eq:constraints1} states that for every object $r$, there can be at most one object on 
its top. Eq.~\eqref{eq:constraints2} states that every
object $i$ should be placed exactly at one place.
Eq.~\eqref{eq:constraints3} states that two objects cannot occupy the same location.
Eq.~\eqref{eq:constraints4} states that even if the objects are at different locations, they still 
cannot have any overlap with each other. In another
word, if $O_i$ placed at $\ell_i$ conflicts with $O_j$ placed at $\ell_j$, then 
only one of them can be
placed. We do not need constrain $s_{ir}$ since `chain stacking' already eliminates
 this overlap issue. 

Enforcing the acyclic property could be hard due to the exponential number of possible cycles
in placing strategies. Expressing all the cycles as constraints is infeasible. 
Therefore, we assume a topological order on stacking: $s_{ir}$ can be 1 only if $O_r$ with
configuration $c_r$ does not have smaller projected 2D area on XY-plane than $O_i$ with
configuration $c_i$. This assumption is reasonable as people usually stack small objects on big
ones in practice. This ensures that the optimal placing strategy is acyclic.

\begin{table*}[!t]
\centering
\caption{{\small Stability results for three baselines and our algorithm using different
features (contact, caging, histograms and all combined) with two different kernels (linear and
quadratic polynomial).
}}\label{tbl:stability}
{\scriptsize
\begin{tabular}{@{}l ccc ccc ccc ccc| ccc}
\whline{1.1pt}
\multirow{2}{*}{}& \multicolumn{3}{c}{dish-racks} &\multicolumn{3}{c}{stemware-holder} &\multicolumn{3}{c}{closet} &\multicolumn{3}{c|}{pen-holder} &\multicolumn{3}{c}{\bf{Average}} \\			  
	& $R_0$ 	& P@5 	& AUC	& $R_0$ 	& P@5 	& AUC	& $R_0$ 	& P@5 	& AUC	& $R_0$ 	& P@5 	& AUC	& $R_0$ 	& P@5 	& AUC\\
\whline{.8pt}
chance	    & 1.5	& 0.70	& 0.50	& 2.0	& 0.48	& 0.50	& 2.0	& 0.47	& 0.50	& 2.1	& 0.44	& 0.50	& 1.9	& 0.52	& 0.50\\
vert	    & 1.3	& 0.79	& 0.74	& 2.0	& 0.70	& 0.77	& 2.5	& 0.63	& 0.71	& 1.0	& 1.00	& 1.00	& 1.7	& 0.78	& 0.80\\
hori	    & 2.1	& 0.22	& 0.48	& 4.3	& 0.35	& 0.54	& 6.0	& 0.03	& 0.36	& 7.2	& 0.00	& 0.38	& 4.9	& 0.15	& 0.44\\
\hline																										
lin-contact	& 1.9	& 0.81	& 0.80	& 2.0	& 0.60	& 0.71	& 1.8	& 0.70	& 0.79	& 1.0	& 0.81	& 0.91	& 1.7	& 0.73	& 0.80\\
lin-caging	& 3.5	& 0.60	& 0.74	& 1.3	& 0.94	& 0.95	& 1.2	& 0.93	& 0.77	& 2.5	& 0.70	& 0.85	& 2.1	& 0.79	& 0.83\\
lin-hist	& 1.4	& 0.93	& 0.93	& 2.3	& 0.70	& 0.85	& 1.0	& 1.00	& 0.99	& 1.0	& 1.00	& 1.00	& 1.4	& 0.91	& 0.94\\
lin-all	    & 1.2	& 0.91	& 0.91	& 2.0	& 0.80	& 0.86	& 1.0	& 1.00	& 1.00	& 1.0	& 1.00	& 1.00	& 1.3	& 0.93	& 0.94\\
poly-contact& 1.2	& 0.95	& 0.88	& 1.0	& 0.80	& 0.85	& 1.2	& 0.93	& 0.93	& 1.2	& 0.91	& 0.88	& 1.1	& 0.90	& 0.89\\
poly-caging	& 2.1	& 0.81	& 0.87	& 2.0	& 0.60	& 0.79	& 1.0	& 0.97	& 0.94	& 2.2	& 0.70	& 0.88	& 1.8	& 0.77	& 0.87\\
poly-hist	& 1.2	& 0.94	& 0.91	& 3.0	& 0.30	& 0.67	& 1.0	& 1.00	& 1.00	& 1.0	& 1.00	& 1.00	& 1.6	& 0.81	& 0.90\\
poly-all	& 1.1	& 0.95	& 0.94	& 1.8	& 0.60	& 0.92	& 1.0	& 1.00	& 1.00	& 1.0	& 1.00	& 1.00	& 1.2	& 0.89	& 0.96\\
\whline{1.1pt}
\end{tabular}
}
\end{table*}

As mentioned before, the search space of placing strategies is too large for applying any
exhaustive search to. 
Several prior works have successfully applied ILP to solve inference in	
Conditional Random Fields or Markov networks, e.g.,
\citealp{roth2005integer, taskar2004learning,
yanover2006linear, globerson2007fixing}. 
Motivated by their approach, we formulate the inference (along with all
constraints) as an ILP problem and then solve it by LP relaxation, which works well in practice.

We use random samples to discretize the continuous space of the location $\ell_i$ and 
configuration $c_i$.\footnote{In our experiments, we randomly
sample $\ell_i$ in about every $10$cm $\times$ $10$cm area. 
An object is rotated every $45$ degree along every dimension, thus generating
 $24$ different configurations.}
We abuse the notation for $S$, $T$ and use the same symbols for representing
sampled variables. We use $s_{ijrtk} \in \{0,1\}$ to represent  
object $O_i$ with the $j$th sampled configuration is placed on  object $O_r$ with the $t$th
configuration at the $k$th location. Similarly, we use $t_{ijrk}$ to represent
placing object $O_i$ in configuration $j$ on top of $E_r$ at location $k$.
Now our problem becomes:
\begin{align}
\scriptstyle
\arg\max_{S, T}& \scriptstyle \sum_{ijrk} \left(t_{ijrk}\log \Psi(t_{ijrk}=1) + (1-t_{ijrk})\log
\Psi(t_{ijrk}=0)\right)\nonumber\\
+\scriptstyle \sum_{ijrtk} &\scriptstyle\left(s_{ijrtk} \log \Psi(s_{ijrtk}=1) + (1-s_{ijrtk}) \log
\Psi(s_{ijrtk}=0)\right) \nonumber\\
\label{eq:lp0}
\scriptstyle\text{s.t. } 
&\scriptstyle
	\sum_{ijk} s_{ijrtk}\leq \sum_{pqk} s_{rtpqk} + \sum_{pk} t_{rtpk}, ~\forall
    r,t; \nonumber\\
& \scriptstyle \sum_{jrtk} s_{ijrtk} + \sum_{jrk} t_{ijrk} = 1, ~\forall i;\nonumber\\
 & \scriptstyle   \sum_{ij}t_{ijrk} \leq 1, ~\forall r,k; \nonumber\\ 
&\scriptstyle
	t_{ijrk} + t_{i'j'rk'} \leq 1, ~\forall t_{ijrk} \text{ overlaps } t_{i'j'rk'}.
\end{align}

While this ILP is provably NP-complete in the general case, 
for some specific cases it can be
solved in polynomial time. For example, 
if all objects have to stack in one
pile, then it reduces to a dynamic programming problem. Or if no stacking is
allowed, then it becomes a maximum matching problem. 
For other general cases, an LP relaxation usually works well in practice.
We use an open-sourced Mixed Integer Linear Programming solver \citep{Berkelaar2004} 
in our implementation. 

\section{Experiments on Placing Multiple Objects}
\label{sec:exp_multi}
In these experiments, the input is the raw point-clouds
of the object(s) and the scene, and our output is a placing strategy
composed of the placing location and orientation for each object. Following this
strategy, we can construct the point-cloud of the scene after placing 
(e.g., Fig.~\ref{fig:placing_def} top-right) and then use path planning to guide the robot
to realize it.

We extensively tested our approach
in different settings to analyze different components of our learning
approach. In particular, we considered
single-object placing, placing in a semantically appropriate area, 
multiple-object single-environment placing, and the end-to-end test of placing in
offices and houses. 
We also tested our approach in robotic experiments, where  
our robots accomplished several placing tasks, such as loading a bookshelf and 
a fridge. 

\subsection{Data} 
\label{sec:exp_multi_data}
We consider 98 objects from 16 categories (such as books, bottles, clothes and
toys shown Table~\ref{tbl:semantic}) and 40 different placing
areas in total (e.g., dish-racks, hanging rod, stemware holder, shelves, etc). The robot observes
objects using its Kinect sensor, and combines point-clouds from 5 views. However, the combined
point-cloud is still not complete due to reflections and self-occlusions, and also
because the object can be observed
only from the top (e.g., see Fig.~\ref{fig:placing_def}). For the environment, only a  single-view
point-cloud is used. We pre-processed the data and segmented the object from its background to reduce the noise
in the point-clouds. For most of the placing areas,\footnote{For some racks and holders that were too thin
to be seen by our sensor, we generated synthetic point-clouds from tri-meshes.} our algorithm took 
real point-clouds. For \textit{all the objects},  our algorithm took only real point-clouds as
input and we did not use any assumed 3D model.

\begin{table*}[tb!]
\centering
\caption{{\small Semantic results for three baselines and our algorithm using different
features (BOW, color, curvature, Eigenvalues, Zernike and all combined) with two different kernels (linear and
quadratic polynomial) on AUC metric.
}}\label{tbl:semantic}
{\scriptsize
\begin{tabular}{@{}l c c c c c c c c c c c c c c c c c@{}}
\toprule%
&
\begin{sideways}\parbox{9mm}{big\\ bottles}\end{sideways}	&
\begin{sideways}\parbox{9mm}{books}\end{sideways}	&
\begin{sideways}\parbox{9mm}{boxes}\end{sideways}	&
\begin{sideways}\parbox{9mm}{clothes}\end{sideways}	&
\begin{sideways}\parbox{9mm}{comp.+ accssry}\end{sideways}	&
\begin{sideways}\parbox{9mm}{hats \& gloves}\end{sideways}	&
\begin{sideways}\parbox{9mm}{dishware}\end{sideways}	&
\begin{sideways}\parbox{9mm}{food}\end{sideways}	&
\begin{sideways}\parbox{9mm}{misc. house.}\end{sideways}	&
\begin{sideways}\parbox{9mm}{martini glasses}\end{sideways}	&
\begin{sideways}\parbox{9mm}{mugs}\end{sideways}	&
\begin{sideways}\parbox{9mm}{shoes}\end{sideways}	&
\begin{sideways}\parbox{9mm}{small\\ bottles}\end{sideways}	&
\begin{sideways}\parbox{9mm}{stationery}\end{sideways}	&
\begin{sideways}\parbox{9mm}{toys}\end{sideways}	&
\begin{sideways}\parbox{9mm}{utensils}\end{sideways}	&
\begin{sideways}\parbox{9mm}{AVG}\end{sideways}	\\
\toprule%
chance & .50	&	.50	&	.50	&	.50	&	.50	&	.50	&	.50	&	.50	&	.50	&	.50	&	.50	&	.50	&	.50	&	.50	&	.50	&	.50 & .50\\    
table & .70	&	.61	&	.59	&	.90	&	.69	&	.78	&	.61	&	1.0	&	.65	&	.32	&	.45	&	.60	&	.64	&	.57	&	.66	&	.50 & .66\\
size & .68	&	.67	&	.67	&	.80	&	.81	&	.80	&	.66	&	1.0	&	.83	&	.46	&	.52	&	.90	&	.70	&	.60	&	.82	&	.51 & .72\\
\midrule
lin-BOW	&	.90	&	1.0	&	.98	&	.08	&	.93	&	.96	&	.86	&	1.0	&	.91	&	.68	&	.89	&	.70	&	.96	&	.86	&	.94	&	.79	&	.85\\
lin-color &	.75	&	1.0	&	1.0	&	.15	&	.96	&	.93	&	.59	&	1.0	&	.80	&	.64	&	.72	&	.45	&	.98	&	.95	&	.86	&	.61	&	.79\\
lin-curv.	&	.85	&	.90	&	.89	&	.78	&	.92	&	.85	&	.62	&	1.0	&	.74	&	.61	&	.67	&	.60	&	.86	&	.82	&	.83	&	.48	&	.79\\
lin-Eigen	&	.67	&	.93	&	.93	&	.40	&	.93	&	.95	&	.53	&	1.0	&	.87	&	.43	&	.47	&	.90	&	.89	&	.89	&	.96	&	.53	&	.79\\
lin-Zern. 	&	.86	&	1.0	&	.91	&	.20	&	.86	&	.89	&	.73	&	.80	&	.83	&	.79	&	.74	&	.65	&	.80	&	.82	&	.84	&	.59	&	.77\\
lin-all	&	.93	&	1.0	&	1.0	&	.60	&	.93	&	.96	&	.84	&	1.0	&	.88	&	.71	&	.87	&	.50	&	.99	&	.89	&	.89	&	.82	&	.85\\
poly-BOW &  .86  &   1.0 &   .89  &   .47  &   .90  &   .97  &   .90  &   1.0 &   .89  &   .82  &   .87  &   .75  &   .96  &   .85  &   .87  &   .82  &   .87\\
poly-all	&	.91	&	1.0	&	1.0	&	.62	&	.90	&	.99	&	.93	&	1.0	&	.89	&	.86	&	.89	&	.75	&	1.0	&	.89	&	.88	&	.84	&	.90\\
\bottomrule
\end{tabular}
}
\end{table*}

\subsection{Placing Single New Objects with Raw Point-Cloud as Input\label{sec:learning_test}}
The purpose of this experiment is, similar to Section~\ref{sec:exp_single}, 
to test our algorithm in placing a \textit{new} object in a scene, but instead 
with raw point-clouds composing the training data.
We test our algorithm on the following four
challenging scenes: 
\begin{enumerate}
\item dish-racks, tested with three dish-racks and six plates;
\item stemware holder, tested with one holder and four martini glass/handled cups;
\item hanging clothes, tested with one wooden rod and six articles of clothing on hangers;
\item cylinder holders, tested with two pen-holder/cup and three stick-shaped objects
(pens, spoons, etc.).  
\end{enumerate}
Since there is only a single placing environment in every scene,
only stability features play a role in this test. We manually labeled $620$ placements 
in total, where the negative examples were chosen randomly. 
Then we used leave-one-out  training/testing 
so that the testing object is always \textit{new} to the algorithm.

We compare our algorithm with three heuristic methods:
\begin{itemize}
\item \textit{Chance}. Valid placements are randomly chosen from the samples. 
\item \textit{Vertical placing}. Placements are chosen randomly from samples 
where the object is vertical (the height is greater than the width). This is relevant 
for cases such as placing plates in dish-racks.
\item \textit{Horizontal placing}. Placements are chosen where the object is 
horizontal (opposed to `vertical').  This applies to cases when the placing area
is somewhat flat, e.g., placing a book on a table.
\end{itemize}
For the placing scenarios considered here, the heuristic based on finding flat surface does not apply at all.
Since we do not know the upright orientation for all the objects and it is not well-defined in some
cases (e.g. an article of clothing on a hanger), 
we do not use the flat-surface-upright rule in these experiments.

Results are shown in Table~\ref{tbl:stability}. We use the same three evaluation metrics as in
Section~\ref{sec:eval_single}: $R_0$, P@5 and AUC.
The top three rows of the table shows our three baselines. Very few of these placing areas are `flat', therefore
the horizontal heuristic fares poorly. The vertical heuristic performs perfectly 
in placing in pen-holder (all vertical orientations would succeed here), but its performance in other cases
is close to chance. All the learning algorithms perform better than all the baseline heuristics in the AUC metric. 

For our learning algorithm, we compared the effect of different stability features as well as
using linear and polynomial kernels. The results show that combining all the
features together give the best performance in all metrics. 
The best result is achieved by polynomial kernel with all the features,
giving an average AUC of $0.96$.

\subsection{Selecting Semantically Preferred Placing Areas}
In this experiment, we test if our algorithm successfully learns the
preference relationship between objects and their placing areas.  Given a single
object and multiple placing areas, we test whether our algorithm correctly
picks out the most suitable placing area for placing that object.
As shown in Table~\ref{tbl:semantic}, we tested 98 objects from 16 categories on 11 different
placing areas: the ground, 3 dish-racks, a utensil organizer, stemware-holder, table,
hanging rod, pen-holder and sink. We exhaustively labeled every pair of object
and area, and used leave-one-out method for training and test.
Again, we build three baselines where the best area is chosen 1) by chance; 2) if it is a table 
(many of objects in our dataset can be placed on table); 3) by its size (whether the
area can cover the object or not).

Our algorithm gives good performance in most object categories except
clothes and shoes. We believe this is because clothes and shoes have a lot of variation making
it harder for our algorithm to learn. Another observation is that the heuristic `table' performs well on clothes.
We found that this is because, in our labeled data set, the table was often labeled as the
second best placing area for clothes after the hanging rod in the closet.  
We tested different semantic features with linear SVM and also compared linear and
polynomial SVM on all features combined. In comparison to the best baseline of 0.72,
we get an average AUC of 0.90 using polynomial SVM with all the features. 


\subsection{Multiple Objects on Single Area}
In this section, we consider placing multiple objects in a very limited space:
14 plates in one dish-rack without any stacking; 17 objects including
books, dishware and boxes on a small table so that stacking is needed; and five articles of clothing 
with hangers on a wooden rod. This experiment evaluates stacking and our LP relaxation.  

\begin{figure*}[t!]
\includegraphics[width=0.33\linewidth]{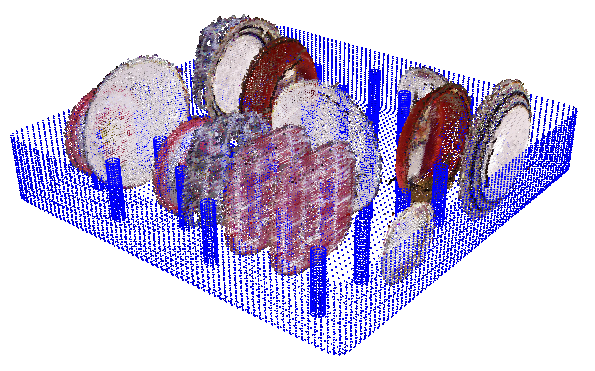}
\includegraphics[width=0.33\linewidth]{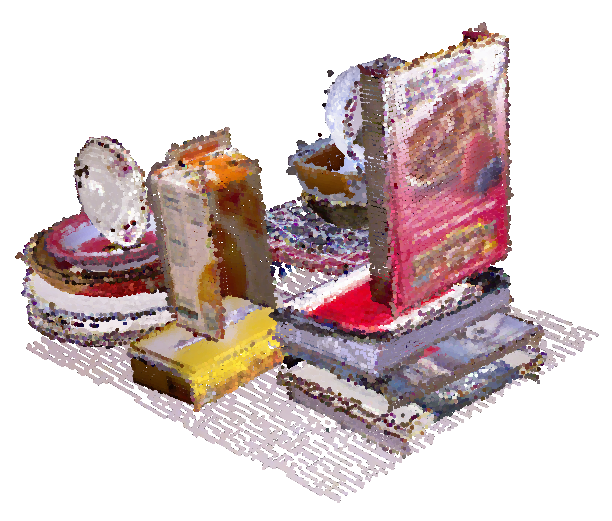}
\includegraphics[width=0.33\linewidth]{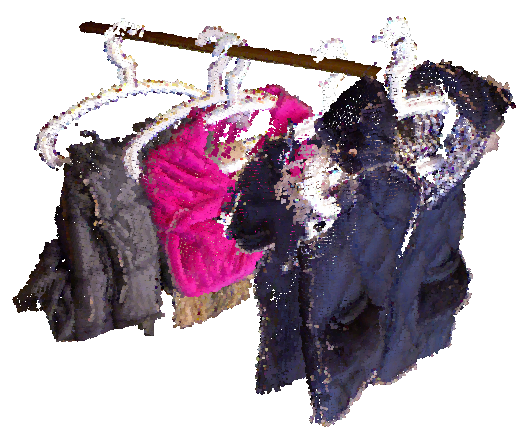}
\caption{{\small Placing multiple objects on a dish-rack (left), a table (middle) and a
hanging rod (right). Most objects are placed correctly, such as the plates vertically in the dish-rack,
books and plates stacked nicely on the table and the hangers with the clothes aligned on the rod.
However, two top-most plates on the table are in wrong configuration and the right-most hanger in
the right figure is off the rod.}}
\label{fig:exp2}
\end{figure*}

The results are shown in Fig.~\ref{fig:exp2}. In the left image, plates are
vertically placed without overlap with spikes and other plates. Most are
aligned in one direction to achieve maximum loading capacity. For the second
task in the middle image, four piles are formed where plates and books are stacked
separately, mostly because of semantic features. Notice that all the dishware, except
the top-most two bowls, is placed horizontally. 
The right image shows all clothes are placed perpendicular upon the rod.
However the first and last hangers are little off the rod due to the outliers in the point-cloud
which are mistakenly treated as part of the object . 
This also indicates that in an actual robotic experiment, the
placement could fail if there is no haptic feedback.

\subsection{Placing in Various Scenes}
In this experiment, we test the overall algorithm with multiple objects being
placed in a scene with multiple placing areas. 
We took point-clouds from 3 different offices and 2 different apartments. 
Placing areas such as tables, cabinets, floor, and drawers were segmented out.
We evaluated the quality of the final placing layout by asking two human subjects 
(one male and one female, not associated with the project) to label placement for each object as
stable or not, semantically correct or not, and also report a qualitative 
metric score on how good the overall placing was (0 to 5 scale).

\begin{figure*}[t!]
\centering
\includegraphics[width=1.0\linewidth]{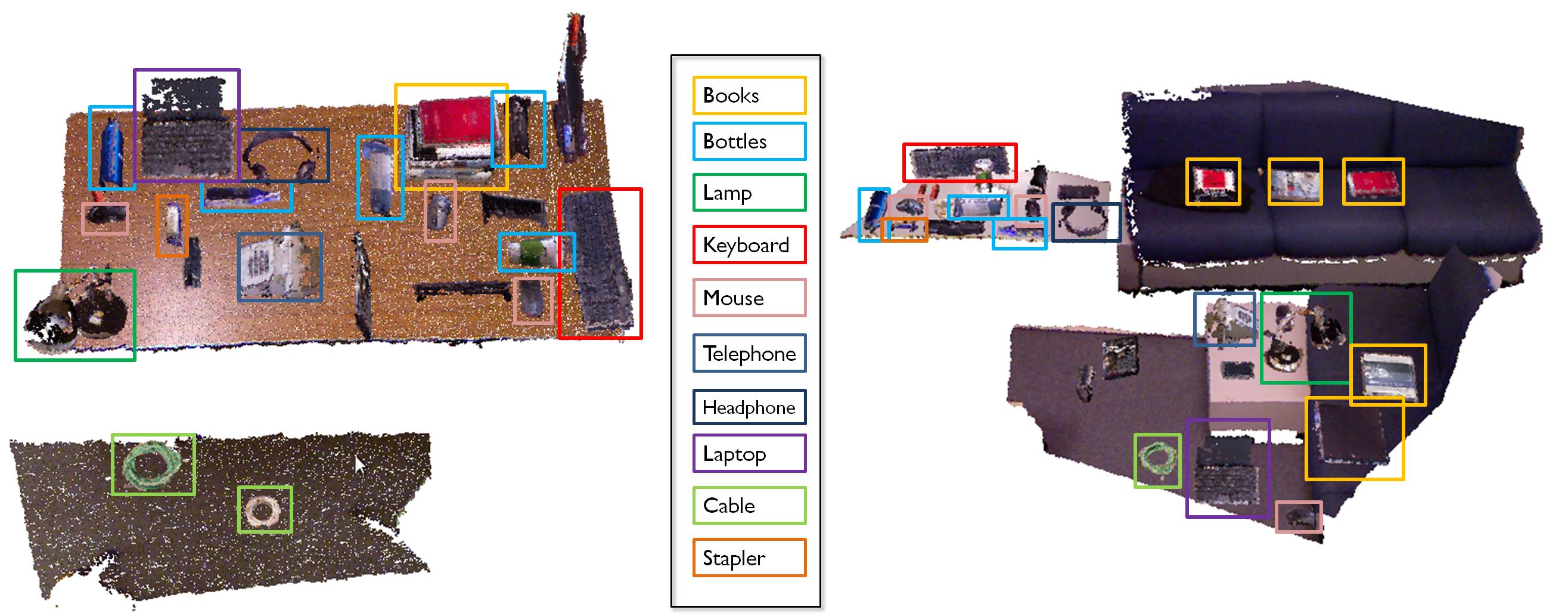}
\caption{{\small Two office scenes after placing, generated by our algorithm. Left scene comprises a table
and ground with several objects in their final predicted placing locations. Right scene comprises
two tables, two couches and ground with several objects placed. }}
\label{fig:office1}
\end{figure*}

Table~\ref{tbl:offline_set} shows the results for different scenes (averaged
over objects and placing areas) and for different algorithms and baselines.
We considered baselines where objects were placed \textit{vertically}, \textit{horizontally} or according to 
configuration \textit{priors} (e.g., flat surface prefers horizontal placing while dish-racks and
holders prefer vertical placing). As no semantic cues were used in the baselines, placing
areas were chosen randomly. 
The results show that with our learning algorithm, the end-to-end performance is
substantially improved under all metrics, compared to the heuristics. 
Fig.~\ref{fig:office1} shows the point-cloud of two offices after placing the objects according 
to the strategy. We have marked some object types in the figure for better visualization.
We can see that books are neatly stacked on the left table,
and horizontally on the couch in the right image. While most objects are placed on the table in the
left scene, some are moved to the ground in the right scene, as the table there is small.

We do not capture certain properties in our algorithm, such as object-object
co-occurrence preferences. Therefore, some placements in our results
could potentially be improved in the future by learning such contextual relations \citep[e.g.,][]{AnandKoppulaNIPS}.
For example, a mouse is typically placed on the side of a keyboard when placed on a table and objects from
the same category are often grouped together. 



\begin{table*}[t!]
\center
\caption{{\small End-to-end test results. In each scene, the number of placing areas and
objects is shown as $(\cdot,\cdot)$. \textbf{St}: \% of
stable placements, \textbf{Co}: \% of semantically correct placements, 
\textbf{Sc}:  average score (0 to 5) over areas.}}\label{tbl:offline_set}
{\scriptsize
\begin{tabular}{@{}l ccc ccc ccc ccc ccc ccc}
\toprule%
& \multicolumn{3}{c}{office-1 (7,29)} & \multicolumn{3}{c}{office-2 (4,29)} & \multicolumn{3}{c}{office-3 (5,29)} & \multicolumn{3}{c}{apt-1 (13,51)} & \multicolumn{3}{c}{apt-2 (8,50)} & \multicolumn{3}{c}{Average} \\ 
& St & Co & Sc & St & Co & Sc & St & Co & Sc & St & Co & Sc & St & Co & Sc & St & Co & Sc\\ 
\toprule
Vert.	&	36	&	60	&	2.4	&	33	&	52	&	2.4	&	52	&	59	&	2.8	&	38	&	41	&	1.9	&	46	&	59	&	3	&	39	&	59	&	2.4	\\
Hori.	&	50	&	62	&	2.9	&	57	&	52	&	2.7	&	69	&	60	&	3.4	&	54	&	50	&	2.5	&	60	&	53	&	2.7	&	57	&	58	&	2.7	\\
Prior	&	45	&	52	&	2.4	&	64	&	59	&	2.8	&	69	&	60	&	3.5	&	44	&	43	&	2.3	&	61	&	55	&	2.7	&	55	&	53	&	2.7	\\
\midrule
Our approach	&	78	&	79	&	4.4	&	83	&	88	&	4.9	&	90	&	81	&	4.5	&	81	&	80	&	3.8	&	80	&	71	&	4.2	&	83	&	82	&	4.4	\\
\bottomrule
\end{tabular}
}
\end{table*}

\begin{figure}
\centering
\includegraphics[width=1.0\linewidth]{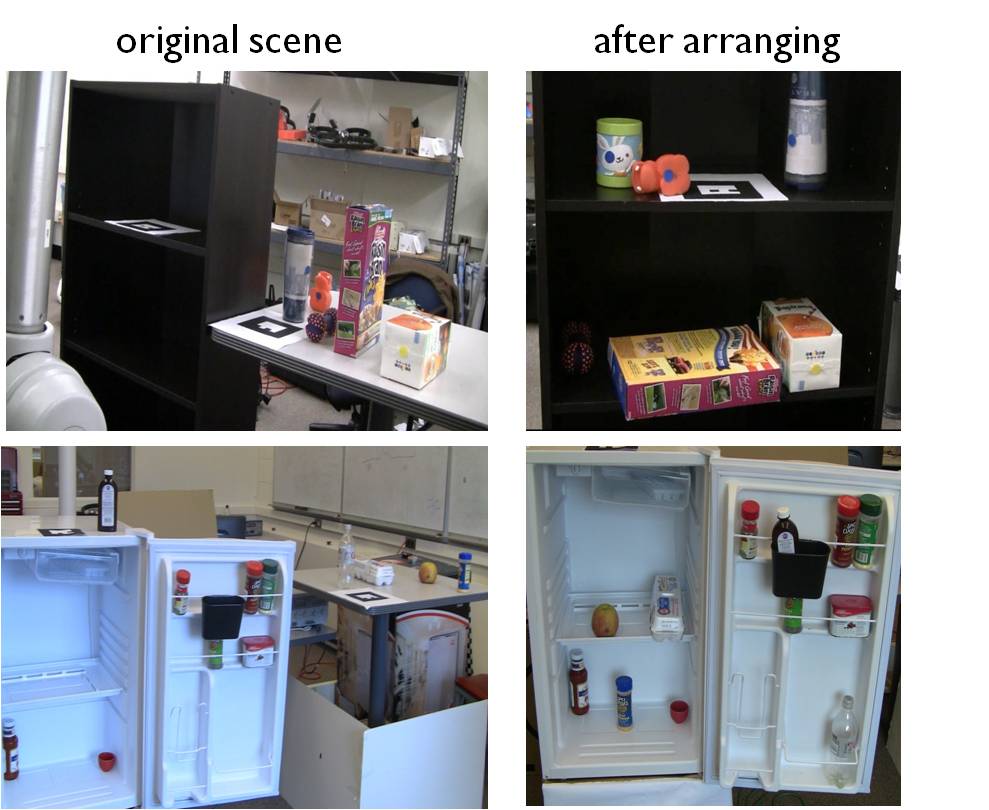} 
\caption{{\small Loading a bookshelf (top) and a fridge (bottom). Snapshots of the scenes before and after the arrangement by our robot POLAR using our learning algorithm.}}
\label{fig:afterclean}
\end{figure}

\begin{figure*}[th!]
\centering
\subfloat[\label{fig:multi_plate} ]{\includegraphics[height=0.19\linewidth]{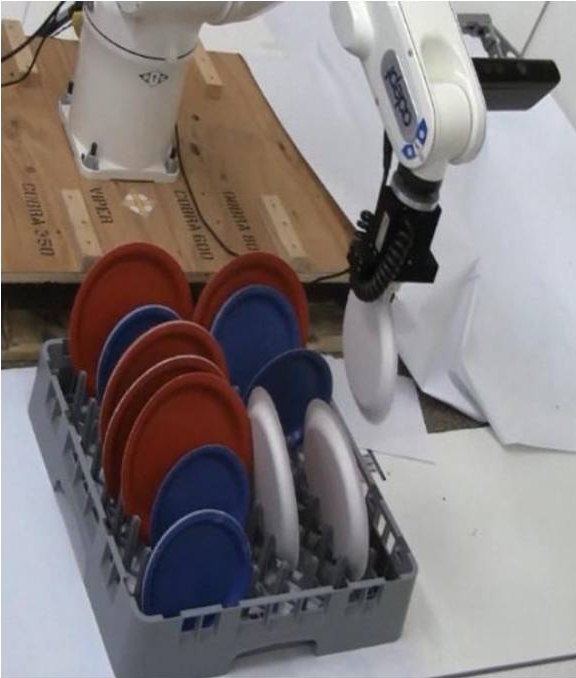}} 
\hspace{.01in}
\subfloat[\label{fig:multiplacing_table1}]{\includegraphics[height=0.19\linewidth]{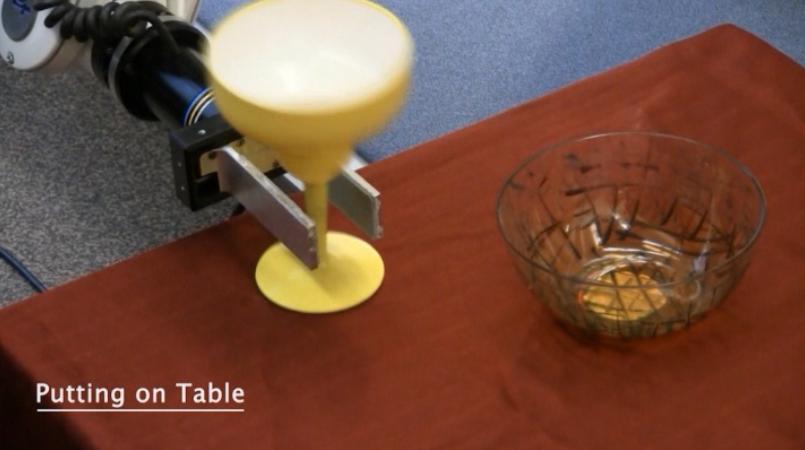}}
\hspace{.01in}
\subfloat[\label{fig:multiplacing_table2}]{\includegraphics[height=0.19\linewidth]{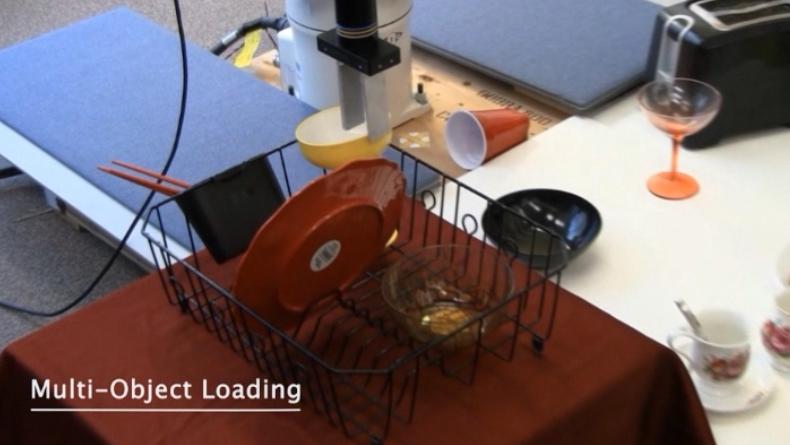}}\\
\subfloat[]{\includegraphics[height=0.21\linewidth]{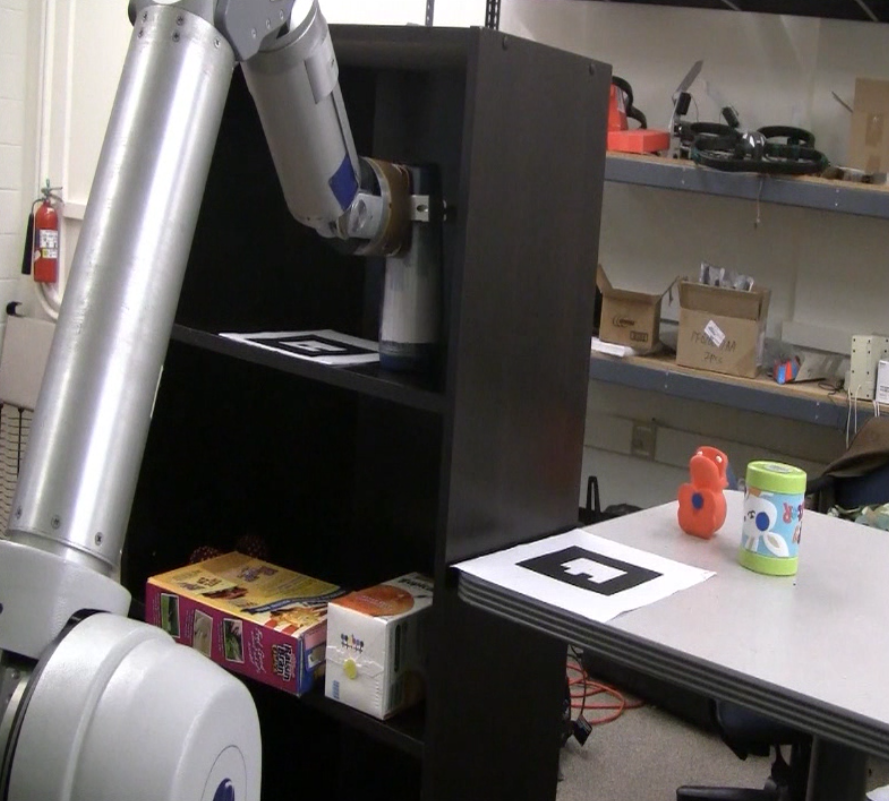}}
\hspace{.01in}
\subfloat[]{\includegraphics[height=0.21\linewidth]{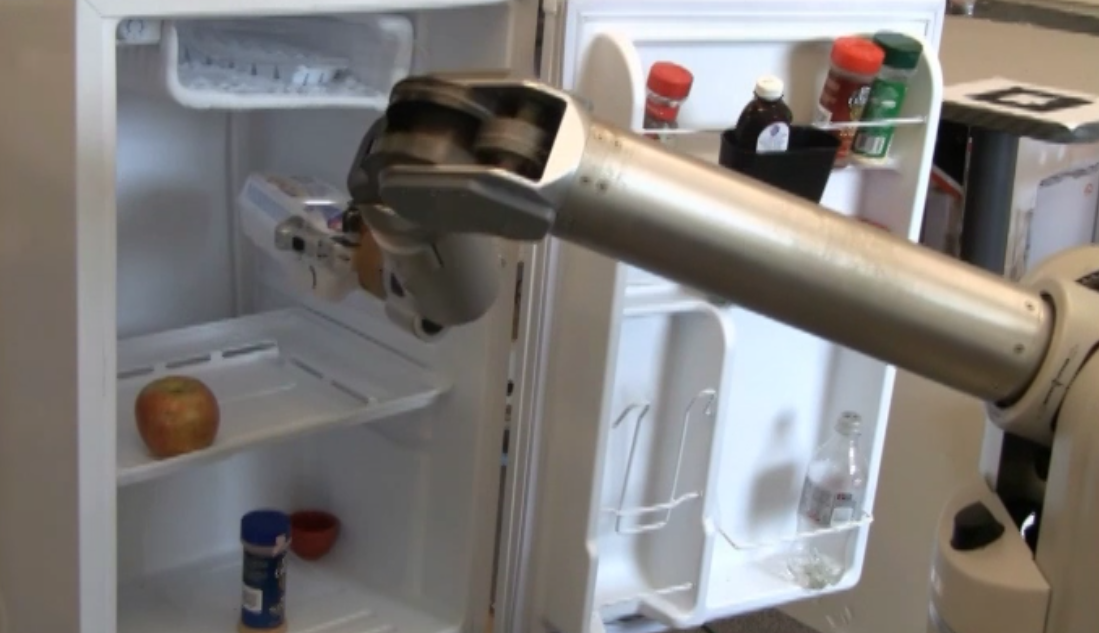}}
\hspace{.01in}
\subfloat[]{\includegraphics[height=0.21\linewidth]{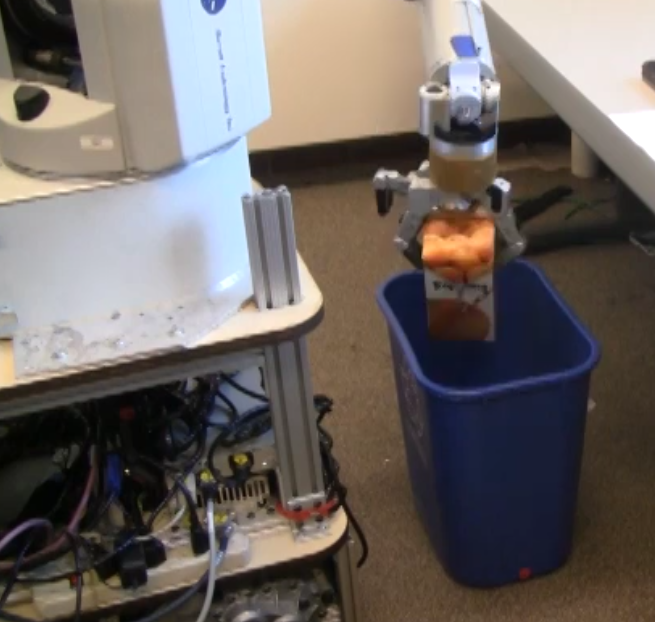}}
\caption{{\small Robots placing multiple objects in different scenes. Top row shows PANDA: (a) loading a dish-rack with
plates, (b) placing different objects on a table, and (c) placing different objects 
in a dish-rack. Bottom row shows POLAR: (d) placing six items
on a bookshelf, (e) loading five items in a fridge, and (d) placing an empty juice box in a recycle 
bin.}}
\label{fig:multiplacing}
\end{figure*}

\subsection{Robotic Experiment}
\label{sec:exp_multi_robotic}

We tested our approach on both our robots, the PANDA and POLAR, with the Kinect sensor. 
We performed the end-to-end experiments as follows. (For quantitative results on placing single objects, see
Section~\ref{sec:exp_single_robotic}.)

In the first experiment, our goal was to place 16 plates in a dish-rack. Once the robot inferred the placing
strategy (an example from the corresponding offline experiment is shown in Fig.~\ref{fig:exp2}-left), 
it placed the plates one-by-one in the dish-rack (Fig.~\ref{fig:multi_plate}). Out of 5 attempts
(i.e., a total of 5x16=80 placements), less than 5\% of placements  failed. This is because the plate 
moved within the gripper  after grasping.

In the second experiment, our aim was to place a cup, a plate and a martini glass together on a table
and then into a dish-rack. 
The objects were originally placed at pre-defined locations and were picked up using the given
grasps. The result is shown in Fig.~\ref{fig:multiplacing_table1} and~\ref{fig:multiplacing_table2}.

In the third experiment, the robot was asked to arrange a room with a bookshelf, a fridge, a
recycle-bin and a whiteboard as potential placing areas. The room contained
14 objects in total, such as bottles, rubber
toys, a cereal box, coffee mug, apple, egg carton, eraser,  cup, juice box, etc. 
Our robot first inferred the placing strategy and then planned a path and placed each one. 
In some cases when the object was far from its placing location, the robot moved to the object,
picked it up, moved to the placing area and then placed it. 
For more details on the system, see Section~\ref{sec:platforms}.
Fig.~\ref{fig:multiplacing} (last row) shows some placing snapshots and the scene before
and after placing is shown in Fig.~\ref{fig:afterclean}. 

Videos of our robot placing these objects are available at:
\texttt{http://pr.cs.cornell.edu/placingobjects}

\section{Conclusion and Future Work}
\label{sec:conclusion}
We considered the problem of placing objects in a scene with multiple
placing areas, in particular, \textbf{where} and \textbf{how} to place each object 
when \textbf{stacking} one on another is allowed. We designed and used appropriate
features that captured both stability and semantic preferences in placing. We 
proposed a graphical model that  encoded these three
properties and certain constraints to force the placing strategy to be feasible.
Inference was expressed as an ILP problem, which was solved using an efficient
LP relaxation. Our extensive experiments demonstrated that our algorithm could
place objects stably and reasonably most of the time. 
In the end-to-end test of placing multiple objects in different scenes, 
our algorithm improved the performance significantly compared to the best baseline. 
Quantitatively, we achieved an average accuracy of 83\% for stable placements and
82\% for semantically meaningful placements.
In our robotic experiments, 
the robots
successfully placed new objects in new placing areas stably with a 82\% 
success rate and 72\% when considering semantically preferred orientations as well.
However, if the robot had seen the object before (and its 3D model was available), 
these numbers were 98\% for both stable as well as preferred placements.
We also used out algorithm on our robots for accomplishing the tasks of loading items 
into dish-racks, a bookshelf, a fridge, etc.

In this work, we have captured only limited contextual information. 
Incorporating such information in future work would allow 
an algorithm to place the objects more meaningfully.
Currently, we have only focussed on using visual and shape features for placing new objects. 
However, there are  several other attributes based on material parameters 
such as surface friction and fragility that
may be relevant for determining good placements.  We believe that
 considering these attributes
 is an interesting direction for future work.  Finally, incorporating haptic and force
sensing while placing would also improve performance, especially when placing
objects in confined spaces such as refrigerators.

{\small
\bibliographystyle{apalike}
\bibliography{references}
}

\end{document}